\documentclass[runningheads]{llncs}

\usepackage{eccv}

\usepackage{eccvabbrv}

\usepackage{graphicx}
\usepackage{booktabs}
\usepackage{multirow} 

\usepackage[accsupp]{axessibility}  

\usepackage{hyperref}
\usepackage{fontawesome}

\usepackage{orcidlink}

\begin{document}

\title{DatasetNeRF: Efficient 3D-aware Data Factory with Generative Radiance Fields} 

\titlerunning{DatasetNeRF}


\author{Yu Chi\inst{1,2}\orcidlink{0009-0005-7821-9808} \and
Fangneng Zhan\inst{2}\orcidlink{0000-0003-1502-6847} \and
Sibo Wu\inst{1}\orcidlink{0009-0009-8140-3971}\and\\
Christian Theobalt\inst{2}\orcidlink{0000-0001-6104-6625}\and
Adam Kortylewski\inst{2,3}\orcidlink{0000-0002-9146-4403}
}

\authorrunning{Y.~Chi et al.}


\institute{Technical University of Munich \and
Max Planck Institute for Informatics \and
University of Freiburg\\
}

\maketitle

\begin{abstract}
  Progress in 3D computer vision tasks demands a huge amount of data, yet annotating multi-view images with 3D-consistent annotations, or point clouds with part segmentation is both time-consuming and challenging. This paper introduces DatasetNeRF, a novel approach capable of generating infinite, high-quality 3D-consistent 2D annotations alongside 3D point cloud segmentations, while utilizing minimal 2D human-labeled annotations. Specifically, we leverage the semantic prior within a 3D generative model to train a semantic decoder, requiring only a handful of fine-grained labeled samples. Once trained, the decoder generalizes across the latent space, enabling the generation of infinite data. The generated data is applicable across various computer vision tasks, including video segmentation and 3D point cloud segmentation in both synthetic and real-world scenarios. Our approach not only surpasses baseline models in segmentation quality, achieving superior 3D-Consistency and segmentation precision on individual images, but also demonstrates versatility by being applicable to both articulated and non-articulated generative models. Furthermore, we explore applications stemming from our approach, such as 3D-aware semantic editing and 3D inversion. Code can be found at \href{https://github.com/GenIntel/DatasetNeRF}{\faGithub/GenIntel/DatasetNeRF}.
\end{abstract}
\begin{figure*}[tb]
  \centering
  \includegraphics[width=0.95\linewidth]{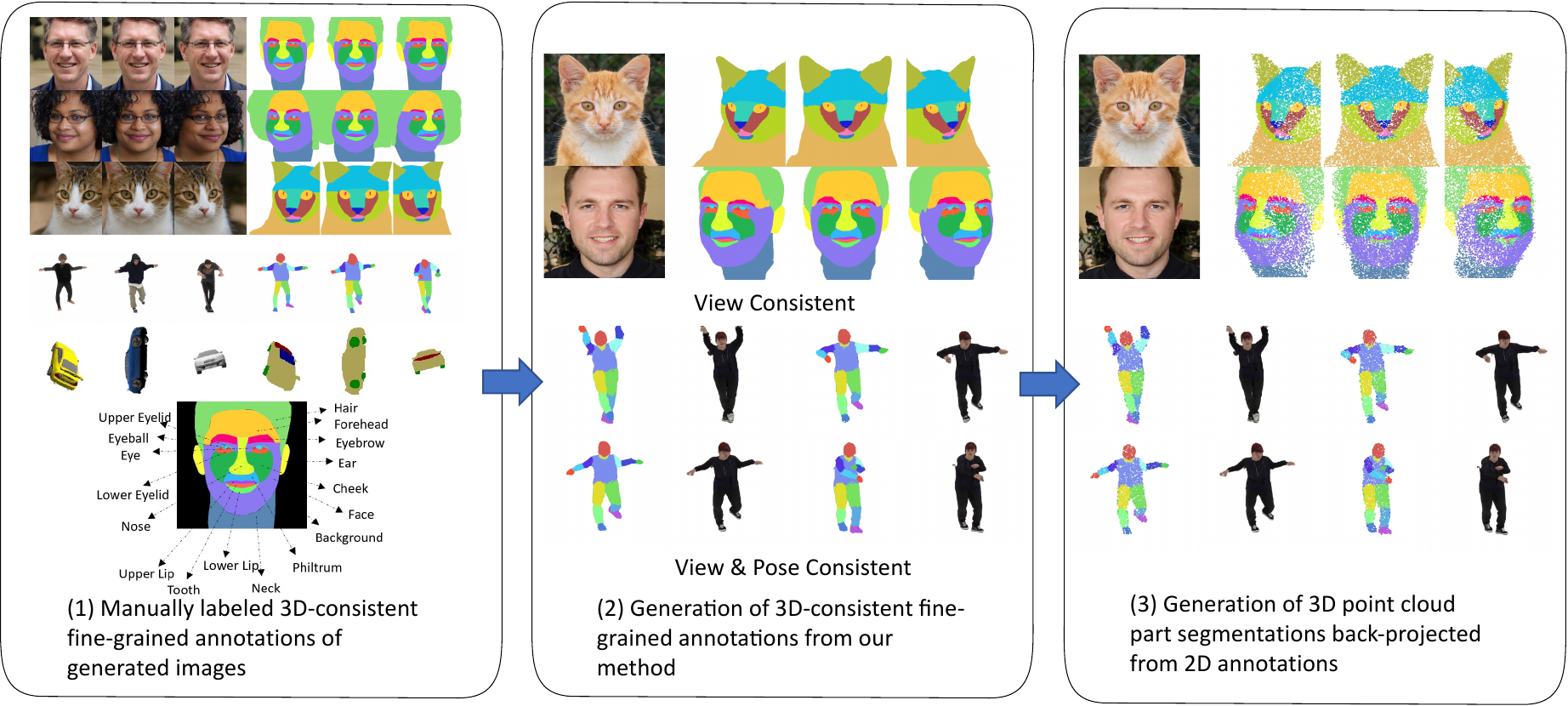}
  \caption{DatasetNeRF Pipeline Overview:
    (1) The manual creation of a small set of multi-view consistent annotations, followed by the training of a semantic segmentation branch using a pretrained 3D GAN backbone. 
    (2) Leveraging the latent space's generalizability to produce an infinite array of 3D-consistent, fine-grained annotations. 
    (3) Employing a depth prior from the 3D GAN backbone to back-project 2D segmentations to 3D point cloud segmentations.
  }
  \label{fig:teaser}
\end{figure*}
\section{Introduction}

In recent years, research on Large-Scale Models or Foundation Models 
has become a prevailing trend. Training these kinds of models demands vast amounts of 2D or 3D labeled data, which entails significant human effort. Based on this limitation, a critical question emerges: How do we efficiently generate a substantial volume of high-quality data-annotation pairs while minimizing human labor? Our paper introduces a method to generate an unlimited supply of high-quality, 3D-aware data by utilizing only a limited set of human-provided 2D annotations.

To efficiently scale datasets, recent approaches \cite{zhang2021datasetgan, baranchuk2021labelefficient, wu2023datasetdm, li2022bigdatasetgan} utilize rich semantic features from 2D generative models as image representations for downstream tasks, such as semantic segmentation. The remarkable representational capacity of generative models facilitates training segmentation models with only a minimal dataset. During inference, a randomly sampled latent code from the generator is capable of producing a corresponding high-quality annotation. This mechanism effectively transforms the generator into an inexhaustible source of data, enabling the creation of extensive datasets with significantly reduced labeling requirements. However, existing methods predominantly focus on 2D generation models, limiting their capability for 3D-aware tasks. Nevertheless, the emergence of geometry-aware 3D Generative Adversarial Networks (GANs) \cite{chan2022efficient, orel2022stylesdf, chan2021pigan}, which decouple latent code and camera pose, offers promising avenues.

In this paper, we introduce \textit{DatasetNeRF}, an efficient 3D-aware data factory based on generative radiance fields. 
Our 3D-aware Data Factory is adept at creating extensive datasets, delivering high-quality, 3D-consistent, fine-grained semantic segmentation, and 3D point cloud part segmentation as shown in Figure \cref{fig:teaser}. This is accomplished by training a semantic branch on a pre-trained 3D GAN, such as EG3D\cite{chan2022efficient}, leveraging the semantic features in the generator's backbone to enhance the feature tri-plane for semantic volumetric rendering. To improve the 3D consistency of our segmentations, we incorporate a density prior from the pre-trained EG3D model into the semantic volumetric rendering process. We further exploit the depth prior from the pre-trained model, efficiently back-projecting the semantic output to obtain 3D point cloud part segmentation. Our approach facilitates easy manipulation of viewpoints, allowing us to render semantically consistent masks across multiple views. By merging the back-projected point cloud part segmentations from different perspectives, we can achieve comprehensive point cloud part segmentation of the entire 3D representation. Remarkably, our process for generating this vast array of 3D-aware data requires only a limited set of 2D data for training.

We evaluate our approach on the AFHQ-cat\cite{choi2020stargan}, FFHQ\cite{karras2019stylebased}, AIST++ dataset\cite{li2021ai} and Nersemble dataset\cite{Kirschstein_2023}. We generate detailed annotations for these datasets, showing our method outperforms existing baselines by enhancing 3D consistency across video sequences and improving segmentation accuracy for single images. Additionally, we demonstrate that our method is also seamlessly compatible with articulated generative radiance fields\cite{bergman2023generative}  on AIST++ dataset. 

 In addition, we qualitatively demonstrate that models trained with our generated dataset can generalize well to real-world scans, such as those in the Nersemble dataset\cite{Kirschstein_2023}.
We also augment the point cloud semantic part segmentation benchmark dataset\cite{Yi16} using our method, with a specific focus on the ShapeNet-Car dataset\cite{chang2015shapenet}. Our work further analyzes potential applications like 3D-aware semantic editing and 3D inversion, demonstrating that the ability to generate infinite 3D-aware data from a limited number of 2D labeled annotations paves the way for numerous 2D and 3D downstream applications.

\section{Related Work}

\subsection{Neural Representations and Rendering}
In recent years, the emergent implicit neural representation offers efficient, memory-conscious, and continuous 3D-aware representations for objects~\cite{park2019deepsdf, atzmon2020sal, michalkiewicz2019implicit, gropp2020implicit} and scenes~\cite{mildenhall2021nerf, sitzmann2020implicit, sitzmann2019scene, zhu2022nice, sucar2021imap, martin2021nerf, muller2022instant} in arbitrary resolution.  By combining implicit neural representation with volume render, NeRF~\cite{mildenhall2021nerf} and its descendants~\cite{garbin2021fastnerf, hedman2021baking, kellnhofer2021neural, lindell2021autoint, liu2020neural, tancik2022block, liu2020dist, ma2022deblur, neff2021donerf, srinivasan2021nerv, zhang2020nerf++, lin2023vision, yang2023contranerf} have yielded promising results for both 3D reconstruction and novel view synthesis applications. Along with image synthesis, the implicit representations are also used to predict semantic maps~\cite{zhi2021place, vora2021nesf, kohli2020semantic}. For example, Semantic-NeRF~\cite{zhi2021place} augments the original NeRF by appending a segmentation renderer. NeSF~\cite{vora2021nesf} learns a semantic-feature grid for semantic maps generation. However, querying properties for each sampled point leads to a low training and inference speed. Considering the pros and cons of explicit representations and implicit representations, recent works~\cite{muller2022instant, chen2022tensorf, bautista2022gaudi, bergman2022generative,deng20233d,zhan2023general} propose hybrid representations to complement each other. In this work, we also use hybrid tri-plane representations for 3D modeling.

\noindent
\subsection{3D Generative Models}
The Generative Adversarial Networks (GANs)~\cite{goodfellow2014generative} have demonstrated remarkable capabilities in generating photorealistic 2D images~\cite{karras2019style, karras2021alias, fei2023masked, karras2020analyzing, zhan2023multimodal}. With this success, some works extended this setting to 3D domain. For instance, PrGANs~\cite{gadelha20173d} and VON~\cite{zhu2018visual} first learn to generate a voxelized 3D shape and then project it to 2D. BlockGAN~\cite{nguyen2020blockgan} learns 3D features but separates one scene into different objects. However, these approaches encounter challenges in achieving photorealistic details due to the limited grid resolutions. 

Recent works~\cite{deng20233d, chan2022efficient, schwarz2020graf, chan2021pigan, gu2021stylenerf, niemeyer2021giraffe, or2022stylesdf, sun2022ide3d} integrated neural implicit representation into GANs to enable 3D-aware image synthesis with multi-view consistency.
Specifically, GRAF~\cite{schwarz2020graf} combines NeRF for scene representation with an adversarial framework to enable the training on unposed images; pi-GAN~\cite{chan2021pigan} operates in a similar setting but makes some differences in network architecture and training strategy; EG3D~\cite{chan2022efficient} learns tri-plane hybrid 3D representation and interprets the aggregated features via volume rendering, ensuring expressive semantic feature extraction and high-resolution geometry-aware image synthesis. While the learned features in generative models are aggregated to generate 3D-aware images, there is still space to harness them for other proposes. 

\noindent
\subsection{Synthetic Dataset Generation}
Traditional dataset synthesis~\cite{dosovitskiy2017carla, puig2018virtualhome, richter2016playing, ros2016synthia} relies on computer simulations for rendering images along with their corresponding labels, which can save annotation cost. However, models trained on such datasets often face challenges in generalizing to real-world datasets due to domain gaps. Unlike traditional methods of dataset synthesis, the use of generative models for dataset synthesis is favored due to their ability to produce a large number of high-quality and diverse images with similar distribution of natural data~\cite{yang2023ai}. The family of generative models is extensive, with GANs~\cite{goodfellow2014generative}, diffusion models~\cite{ho2020denoising}, and NeRF~\cite{mildenhall2021nerf} having achieved notable success in image synthesis. Specifically, many works leverage the rich semantic information learned by GANs to manipulate images~\cite{shen2020interpreting, ling2021editgan, alaluf2022hyperstyle}. Diffusion models benefit from a stationary training objective and demonstrate decent scalability, enabling the generation of high-quality images~\cite{dhariwal2021diffusion}. NeRF, as a recent and emerging generative model, has received widespread acclaim for maintaining multi-view consistency. Additionally, the capability of generative models to learn rich semantic information allows such methods to learn to generate new data and labels using only a few manually annotated images~\cite{nguyen2023dataset, li2022bigdatasetgan, zhang2021datasetgan, wu2023datasetdm}. For instance, DatasetGAN~\cite{zhang2021datasetgan} leverages StyleGAN~\cite{karras2019style} as an image generator and synthesize accurate semantic labels with a few human labeled data. Nevertheless, these previous efforts focused on generating semantic maps for 2D datasets. 
\begin{figure*}[ht]
    \centering
    \includegraphics[width=\textwidth]{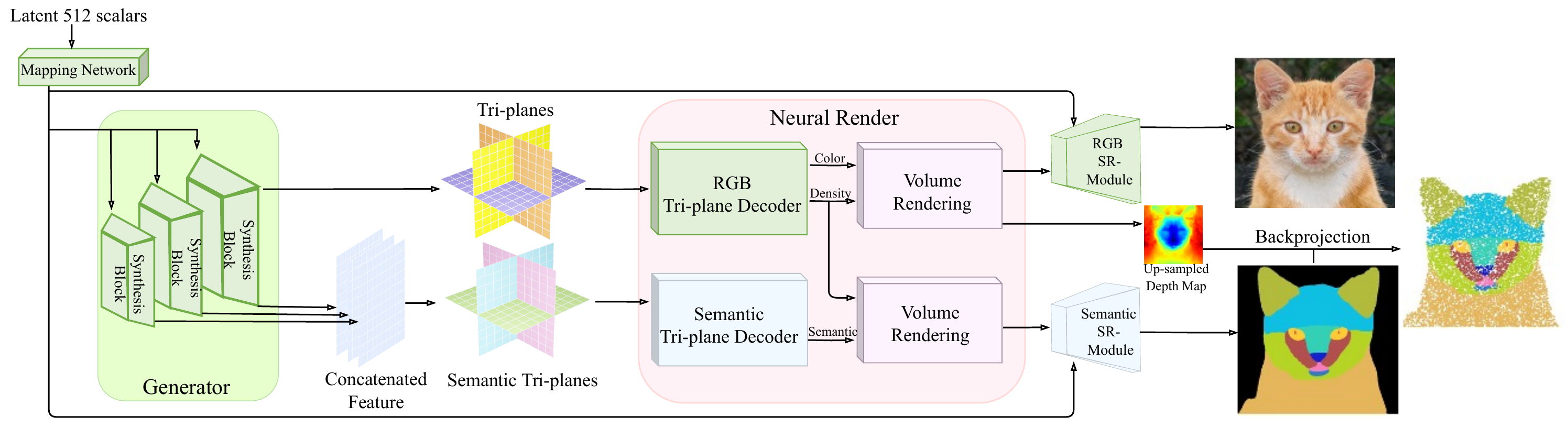}
    \caption{Overall Architecture of DatasetNeRF. The DatasetNeRF architecture unifies a pretrained EG3D model with a semantic segmentation branch, comprising an enhanced semantic tri-plane, a semantic feature decoder, and a semantic super-resolution module.  The semantic feature tri-plane is constructed by reshaping the concatenated outputs from all synthesis blocks of the EG3D generator. The semantic feature decoder   interprets aggregated features from semantic tri-plane into a 32-channel semantic feature for every point. The semantic feature map is rendered by semantic volumetric rendering. We incorporate a density prior from the pretrained RGB decoder during the rendering process to enhance 3D consistency. The semantic super-resolution module upscales and refines the rendered semantic feature map into the final semantic output. The combination of the semantic mask output and the upsampled depth map from the pretrained EG3D model enables an efficient process for back-projecting the semantic mask, thereby facilitating the generation of point cloud part segmentation.}
    
    \label{fig:fullmodelarchitecture}
\end{figure*}
\section{Method}

We introduce \textit{DatasetNeRF}, a framework designed to generate an extensive range of 3D-aware data. It efficiently produces fine-grained, multi-view consistent annotations and detailed 3D point cloud part segmentations from a limited collection of human-annotated 2D training images.

To address the challenge of generating a varied 3D-aware dataset, we employ a 3D GAN generator as the foundational architecture of our framework. We augment this 3D GAN with a semantic segmentation branch, enabling the production of precise annotations across diverse 3D viewpoints as well as detailed 3D point cloud part segmentations. \cref{fig:fullmodelarchitecture} provides a comprehensive visualization of the entire model architecture. For a more in-depth understanding of the different components of our framework, we delineate the specific backbones used for various tasks in \cref{subsec:3dgan}. Subsequently, in \cref{subsec:semanticbranch}, we elaborate on the methodology employed to train the semantic segmentation branch. 
\begin{figure}[h]
    \centering
    \includegraphics[scale=0.3]{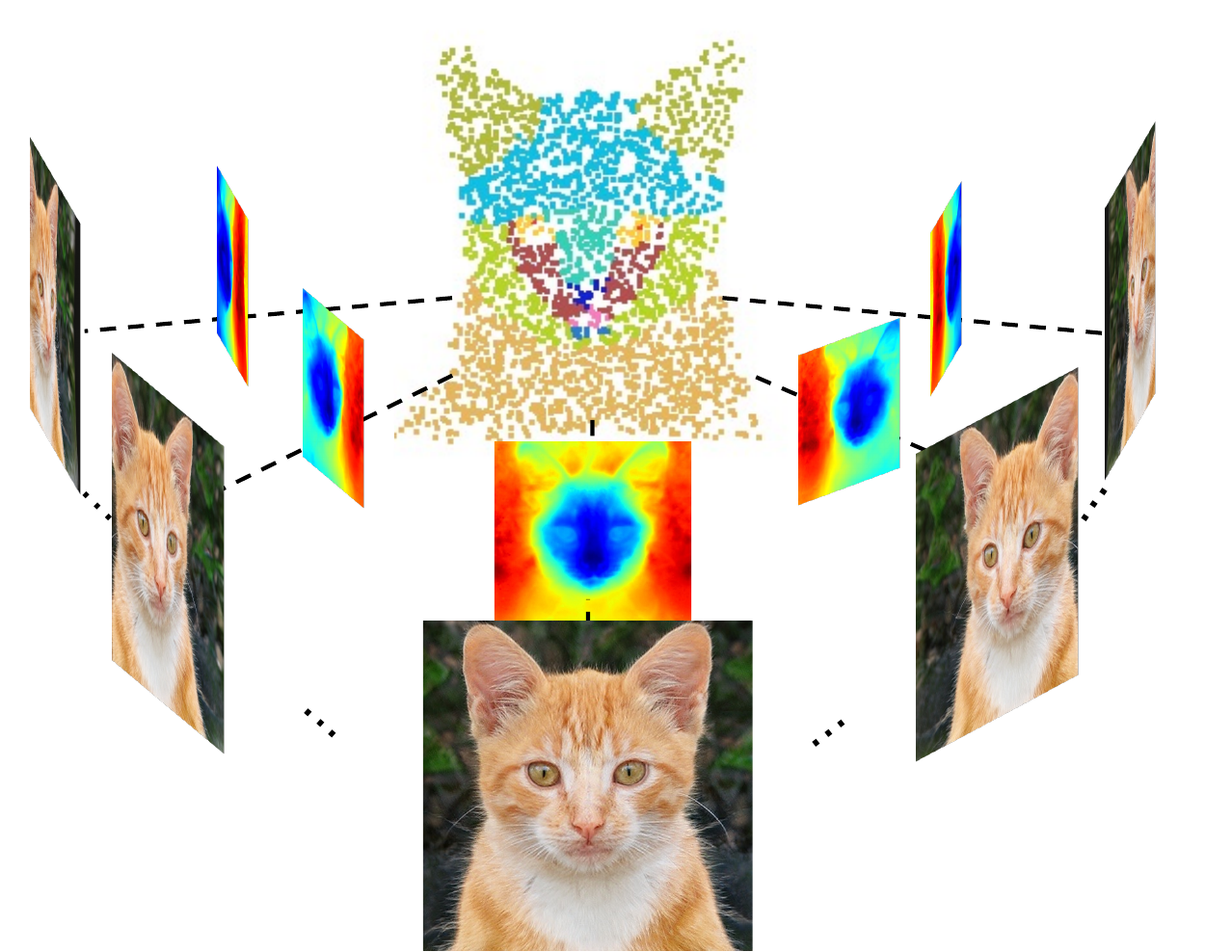}
    \caption{The illustration of multi-view point cloud fusion.}
    \label{fig:pointcloudfusion}
\end{figure}
In \cref{subsec:3dfactory}, we provide a detailed presentation of both the generation process and the resulting 3D-aware data within the DatasetNeRF framework.
\subsection{3D GAN Generator Backbone}
\label{subsec:3dganbackbone}

We take EG3D~\cite{chan2022efficient} as our backbone model, which introduces a tri-plane architecture for efficient neural rendering at reduced resolutions. This tri-plane consists of reshaped feature representations derived from the output of the generator. To enhance the representational power of the triplane in our work, we take all feature maps \(\{S^{0}, S^{1}, \ldots, S^{k}\}\) from the the output of consecutive synthesis block of the pretrained generator. These features are upsampled to match the highest output resolution, and subsequently concatenated into a feature tensor with dimensions \([N, N, C]\). 
Following this, we reshape the concatenated feature tensor into an augmented tri-plane format, similar to that of EG3D\cite{chan2022efficient}, to facilitate our semantic neural rendering pipeline. This tri-plane format, distinct from other work\cite{chan2022efficient}, represents a key innovation in our work. Its significance and impact are further validated through an ablation study detailed later in the text. Our enhanced tri-plane serves as the semantic feature volume for rendering semantically-rich features within our semantic segmentation branch, enabling accurate depiction of complex structures in images.

Notably, our methodology also exhibits compatibility with a range of 3D GAN architectures, whether articulated or inarticulated. This adaptability underscores the robustness of our approach in generating 3D-consistent segmentation. It facilitates not only multi-view consistency but also pose-consistency in segmentations, further proving the utility of our approach across diverse tasks.
\label{subsec:3dgan}

\subsection{Semantic Segmentation Branch Training}
\label{subsec:semanticbranch}

We query any 3D position \(x\) within our semantic tri-plane with enhanced 
format by projecting it onto each of the three feature planes, obtaining the respective feature vectors \((F_{xy}, F_{xz}, F_{yz})\) through bi-linear interpolation. These vectors are then aggregated via summation. This aggregated feature serves as the input to the subsequent semantic decoder, which outputs a 32-channel semantic feature. To harness the 3D consistency inherent in the pretrained EG3D model, we re-use the same density \(\sigma\) as the pretrained RGB decoder at the equivalent tri-plane point. For a majority of our experiments, the semantic feature map is rendered at a resolution of \(128^2\). Through semantic volumetric rendering, we derive a raw semantic map \(\hat{I}_{s} \in \mathbf{R}^{128 \times 128 \times C}\) and a semantic feature map \(\hat{I}_{\phi} \in \mathbf{R}^{128 \times 128 \times 32}\). Subsequently, a semantic super-resolution module \(U_{s}\) is utilized to refine the semantic map into a high-resolution segmentation \(\hat{I}^{+}_{s} \in \mathbf{R}^{512 \times 512 \times C}\):

\begin{equation}
\hat{I}^{+}_{s} = U_{s}(\hat{I}_{s}, \hat{I}_{\phi}).
\end{equation}
For a given ground-truth viewpoint \(P\) and corresponding latent code \(\mathbf{z}\), we compare the ground-truth semantic mask \(I_{s}\) with our model's output semantic mask using cross-entropy loss, mathematically represented as:

\begin{equation}
\mathcal{L}_{\text{CE}}(I_{s}, \hat{I}^{+}_{s}) = -\sum_{c=1}^{C} I_{s,c} \log(\hat{I}^{+}_{s,c}),
\end{equation}
where \(I_{s,c}\) is the binary indicator of the ground-truth class label for class \(c\) and \(\hat{I}^{+}_{s,c}\) is the predicted probability of class \(c\) for each pixel in the final output.

\subsection{DatasetNeRF as 3D-Aware Data Factory}
\label{subsec:3dfactory}
\begin{figure*}[hb]
  \centering
  \begin{minipage}[b]{0.49\textwidth}
    \includegraphics[width=\linewidth]{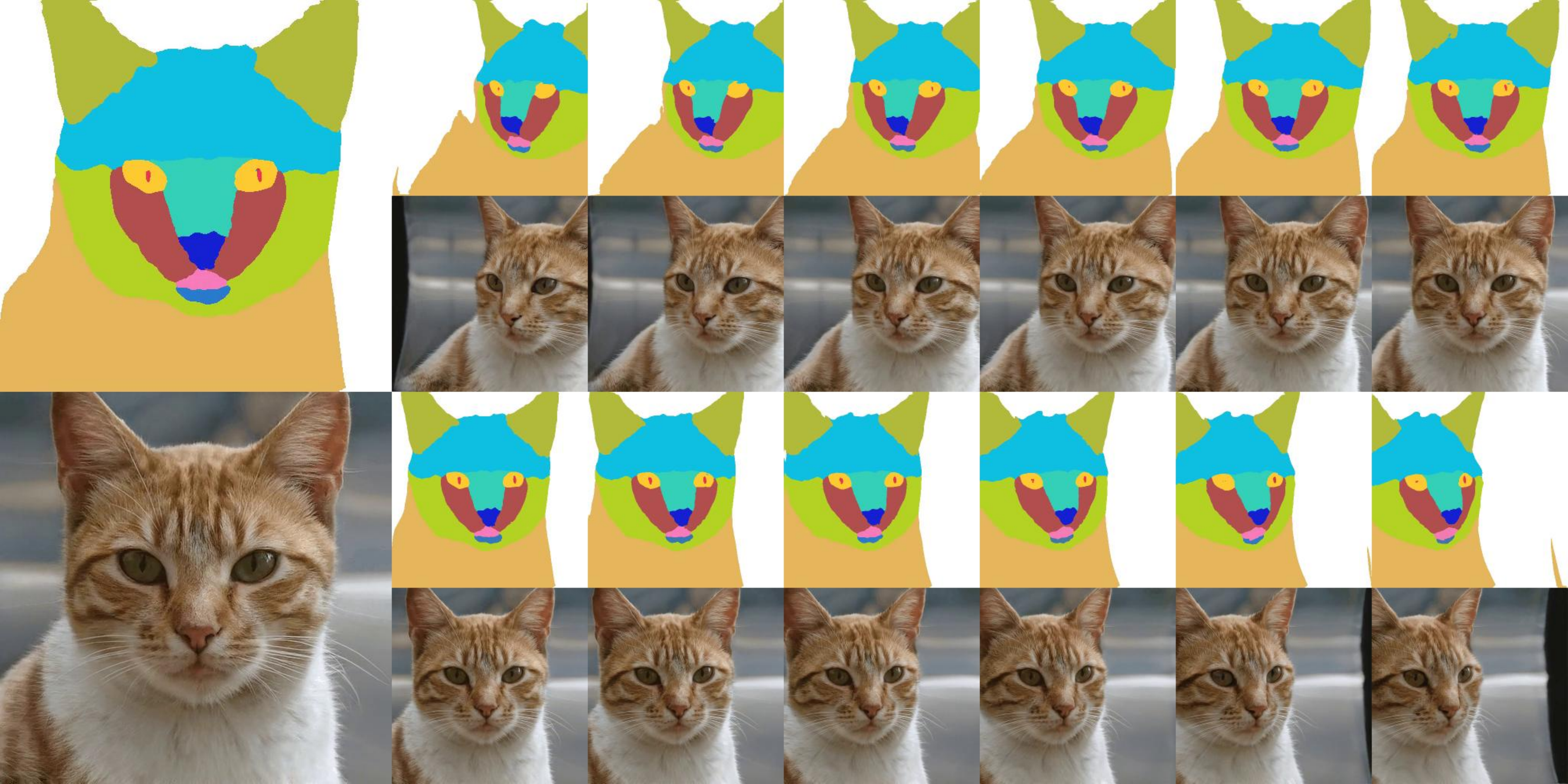}
  \end{minipage}
  \begin{minipage}[b]{0.49\textwidth}
    \includegraphics[width=\linewidth]{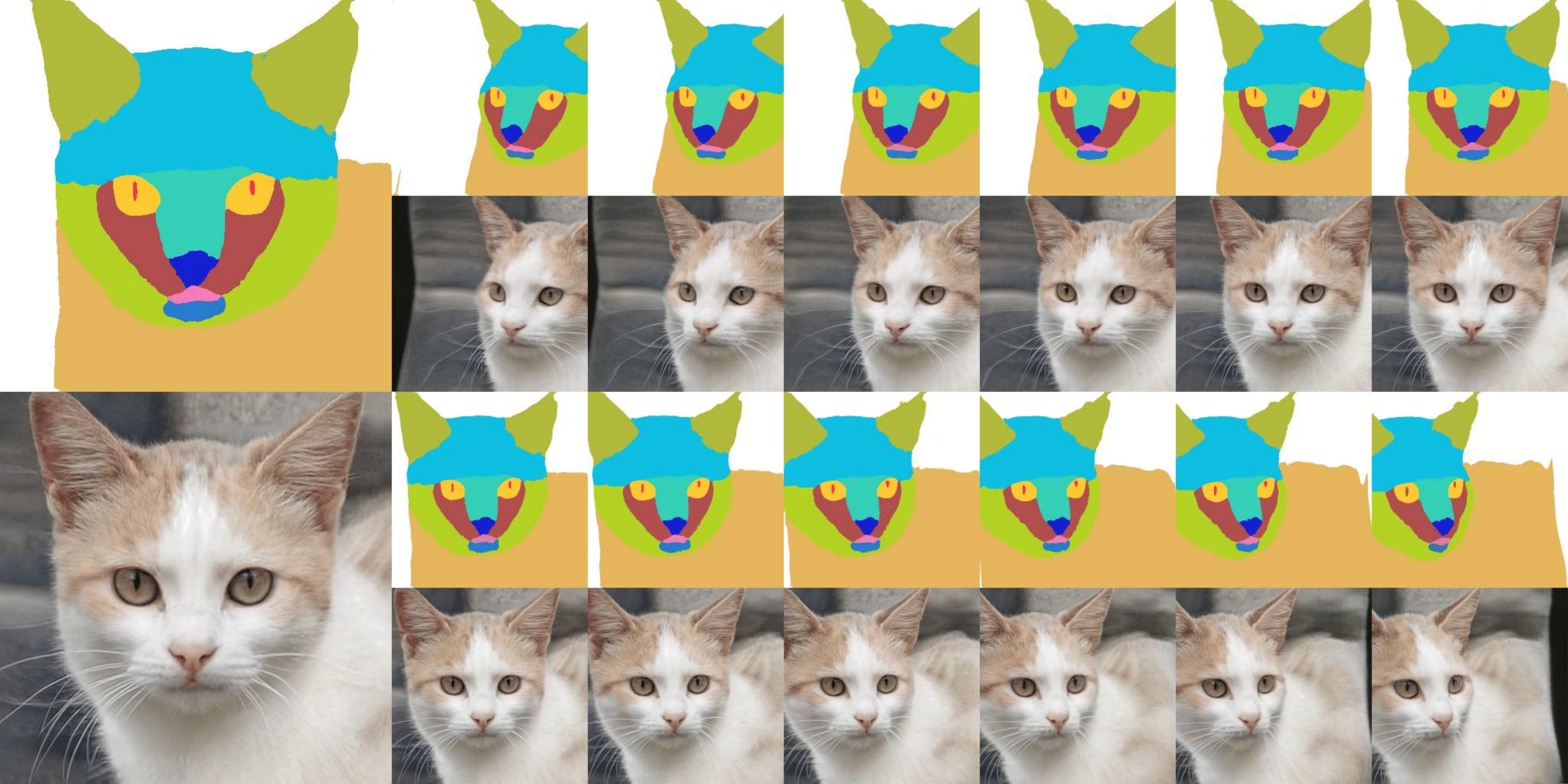}
  \end{minipage}

  \begin{minipage}[b]{0.49\textwidth}
    \includegraphics[width=\linewidth]{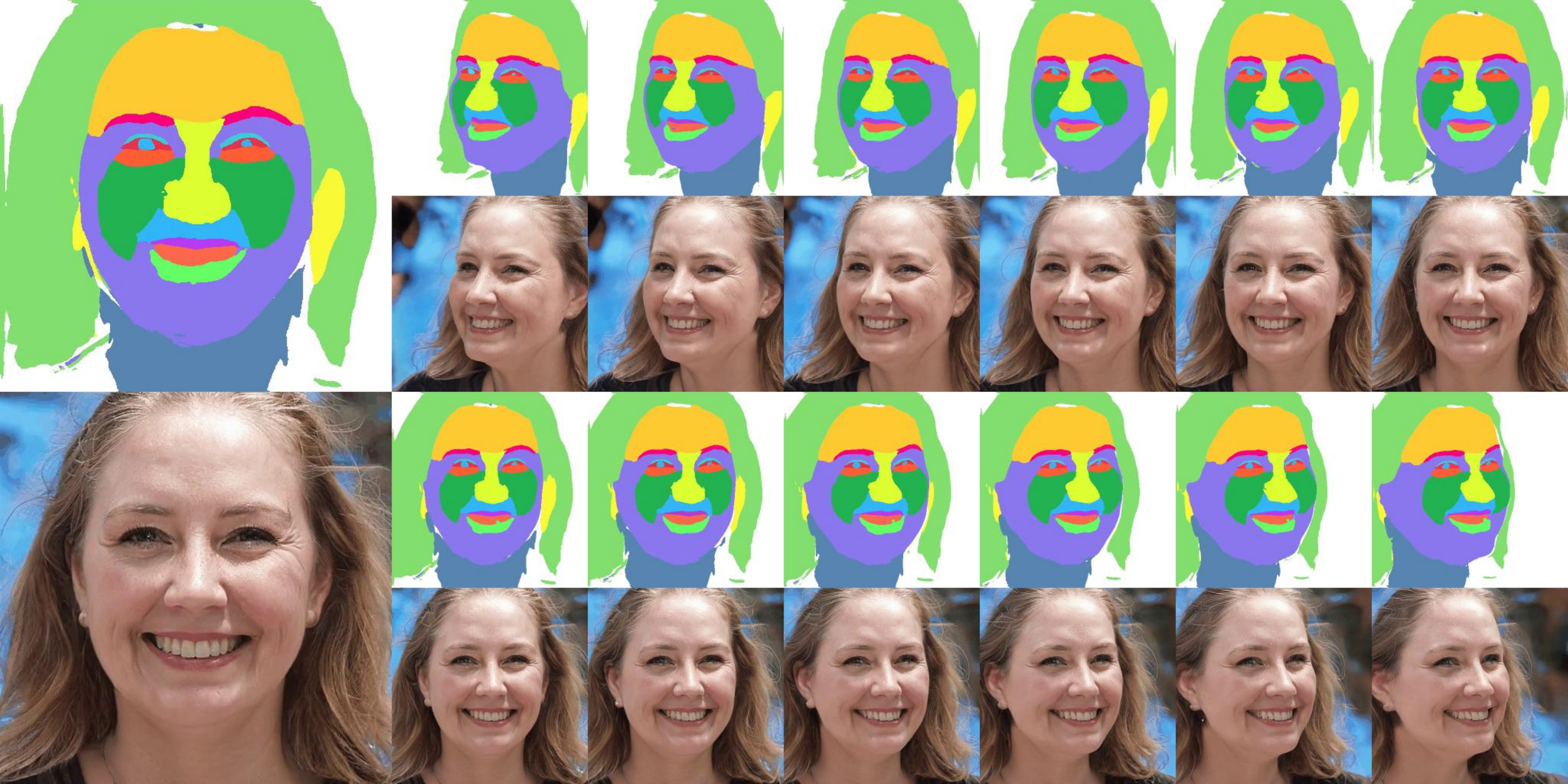}
  \end{minipage}
  \begin{minipage}[b]{0.49\textwidth}
    \includegraphics[width=\linewidth]{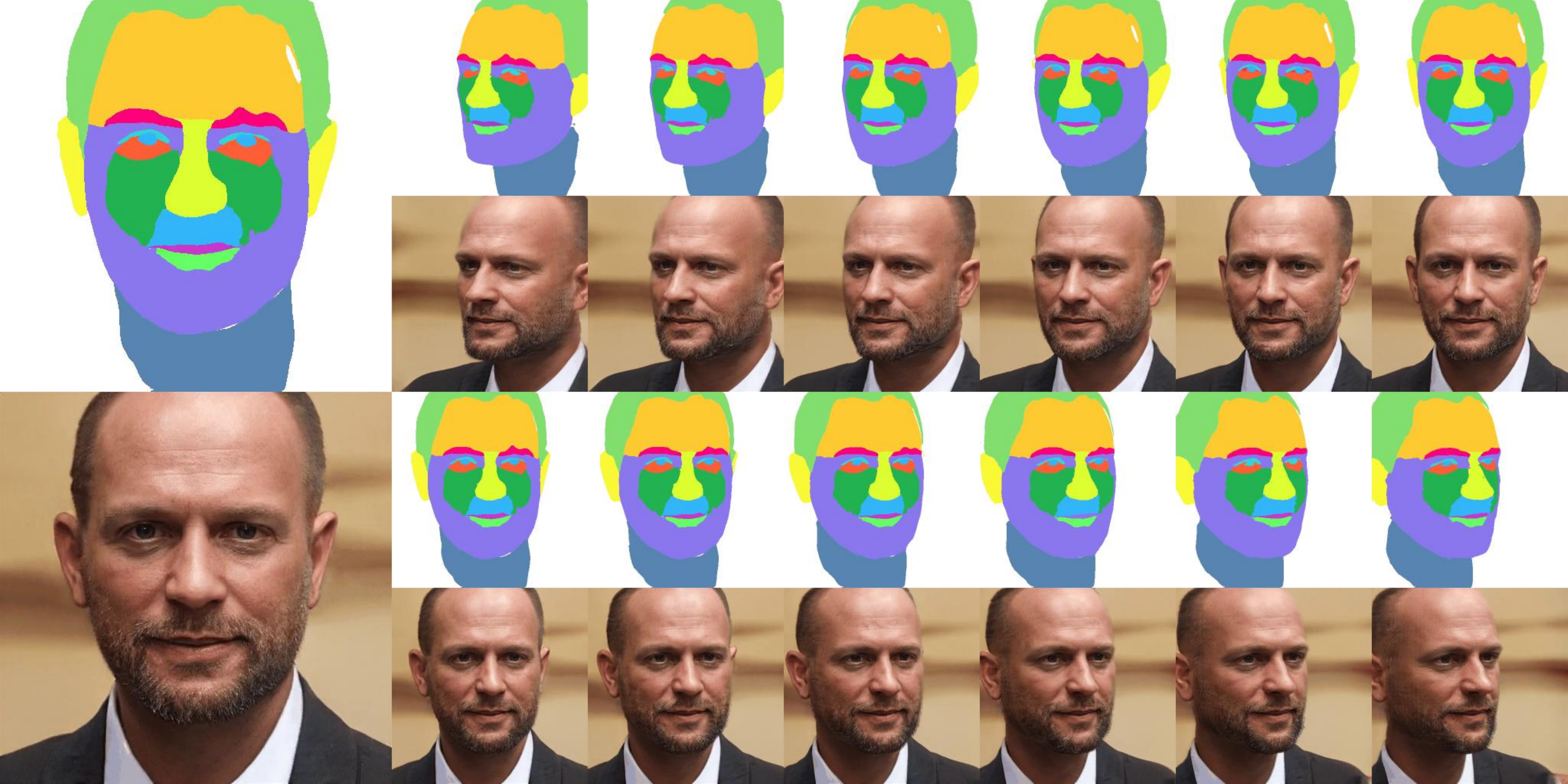}
  \end{minipage}

  \begin{minipage}[b]{0.49\textwidth}
    \includegraphics[width=\linewidth]{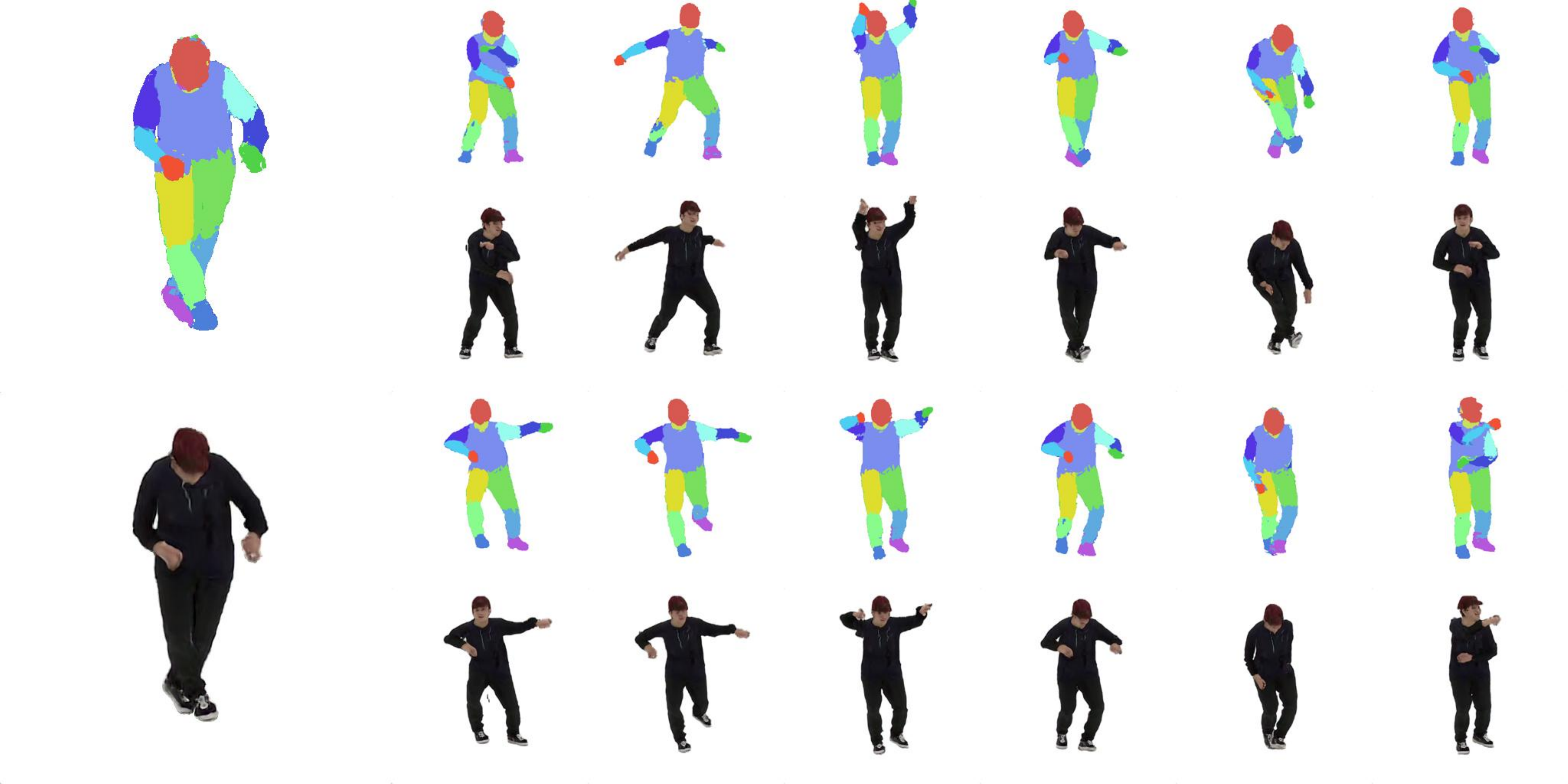}
  \end{minipage}
  \begin{minipage}[b]{0.245\textwidth}
    \includegraphics[width=\linewidth]{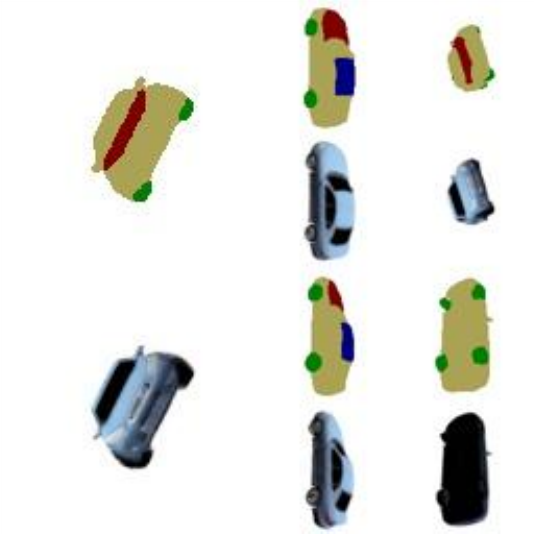}
    
  \end{minipage}
\begin{minipage}[b]{0.245\textwidth}
    \includegraphics[width=\linewidth]{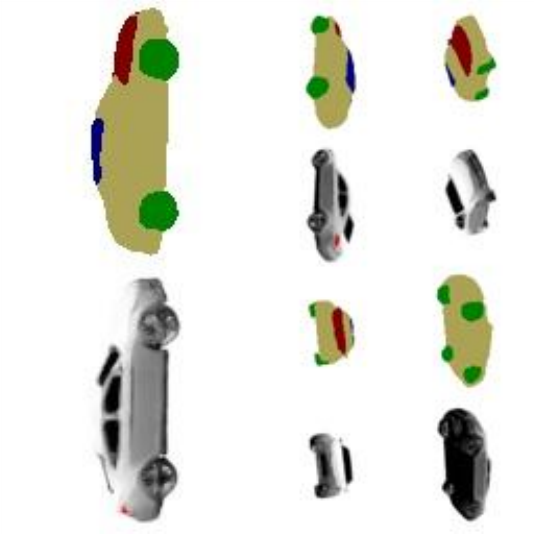}
  \end{minipage}

  \caption{{Examples of synthesized image-annotation pairs from 3D-aware data factoty. } }
  \label{fig:3d_aware_factory}
\end{figure*}

\noindent\textbf{DatasetNeRF as Multi-view Consistent Segmentations Factory.} Empowered by the geometric priors derived from 3D GAN, our DatasetNeRF naturally specializes in generating segmentations that maintain consistency across multiple viewpoints. Once trained, the model adeptly produces high-quality semantic segmentations from a randomly sampled latent code paired with any given pose. The generated multi-view consistent images are illustrated in \cref{fig:3d_aware_factory}. The easy generation of fine-grained, multi-view consistent annotations markedly diminishes the need for human effort.

\noindent\textbf{DatasetNeRF as 3D Point Cloud Segmentation Factory.} 
Initially, we render a depth map using the pretrained RGB branch. The depth maps are generated via volumetric ray marching. This method computes depth by aggregating weighted averages of individual depths along each ray. The depth map is upsampled to align with the dimensions of the semantic mask, allowing the semantic mask to be back-projected into 3D space. The final point cloud of the object is formed by merging back-projected semantic maps from various viewpoints, shown in \cref{fig:pointcloudfusion}. The efficient acquisition of fine-grained 3D point cloud segmentation significantly reduces the amount of human effort required. The visualization of point cloud part segmentation is illustrated in \cref{fig:teaser} (3).

\noindent\textbf{Extension to Articulated Generative Radiance Field.} 
We showcase how our method can also be applied to articulated generative radiance field. Instead of using EG3D, we adopt the generator of GNARF\cite{bergman2022generative} as our backbone. GNARF\cite{bergman2022generative} introduces an efficient neural representation for articulated objects, including bodies and heads, that combines the tri-plane representation with an explicit feature deformation guided by a template shape. The semantic branch is trained on top of the deformation-aware feature tri-plane. The training set contains 150 annotations from 30 different human samples and 60 different training poses. As shown in \cref{fig:3d_aware_factory}, the result can be well generalized on novel human poses.

\section{A Small Dataset with Human Annotations}

Our method necessitates a small dataset with annotation. To this end, we employ our backbone model to synthesize a small number of images, followed by a professional annotator for fine-grained annotation. Our fine-grained annotation protocol was applied to AFHQ-Cat~\cite{choi2020stargan}, FFHQ~\cite{karras2019stylebased}, and AIST++~\cite{li2021ai}, with a simplified scheme utilized for ShapeNet-Car~\cite{chang2015shapenet}.

\noindent\textbf{Annotation Details}
For the training set, we crafted 90 fine-grained annotations for each of the AFHQ-Cat~\cite{choi2020stargan} and FFHQ~\cite{karras2019stylebased} datasets. These encompass 30 distinct subjects with three different views for each subject. The angular disposition for both training and testing spans from $-\frac{\pi}{6}$ to $\frac{\pi}{6}$ relative to the frontal view, holding all other degrees of freedom constant. Consequently, each subject is depicted in a frontal stance, accompanied by both leftward and rightward poses. The AIST++ dataset~\cite{li2021ai} posed a greater challenge due to the diversity of human poses, prompting us to generate 150 annotations for 60 disparate poses across various human subjects. For the ShapeNet-Car dataset, our efforts yielded 90 annotations from 30 distinct samples, each from a unique viewpoint. The annotations for ShapeNet-Car, identifying parts like \textit{hood, roof, wheels, other}, align with the standard labels used in the point cloud part segmentation benchmark\cite{Yi16}. The manually created dataset is visualized in \cref{fig:teaser}(1).

\begin{table*}[htbp]
\caption{{Comparison of different approaches on AFHQ-Cat and FFHQ datasets. In the table, ``Ind.” refers to 90 unique individual image annotations representing arbitrary poses within the training range, and ``Vid.” denotes a video annotation comprising 61 frames.}}
    \centering
    
    \begin{tabular}{lccccccccc}
        \toprule
        \multirow{2}{*}{Methods}
         & \multicolumn{2}{c}{AFHQ-Cat Ind.} & \multicolumn{2}{c}{AFHQ-Cat Vid.} & \multicolumn{2}{c}{FFHQ Ind.} & \multicolumn{2}{c}{FFHQ Vid.} \\
        \cmidrule(lr){2-3} \cmidrule(lr){4-5} \cmidrule(lr){6-7} \cmidrule(lr){8-9}
        & mIoU & Acc. & mIoU & Acc. & mIoU & Acc. & mIoU & Acc. \\
        \midrule
        Transfer Learning & 0.2995 & 0.6358 & 0.2605 & 0.5766 & 0.4083 & 0.7964 & 0.3895 & 0.8611 \\
        DatasetGAN & 0.5381 & 0.8464 & 0.5971 & 0.8625 & \textbf{0.6317} & 0.8881 & 0.6390 & 0.9259 \\
        DatasetNeRF & \textbf{0.6057} & \textbf{0.8798} & \textbf{0.6756} & \textbf{0.9253} & 0.6200 & \textbf{0.8996} & \textbf{0.6561} & \textbf{0.9278} \\
        \bottomrule
    \end{tabular}
    \label{tab:full_comparison}
\end{table*}

\section{Experiments}
\label{sec:exp}
We conduct extensive experiments with our approach. First, we assess the 2D part segmentation performance across two distinct object categories: cat and human faces. Furthermore, we demonstrate the efficacy of our method in generating 3D point cloud part segmentations for both cat faces and ShapeNet-Cars. Finally, we show a variety of 3D applications based on GAN inversion~\cite{zhu2018generative}.

\subsection{2D Part Segmentation}
\label{sec:2dpartseg}
\noindent\textbf{2D Part Segmentation Network.} Consistent with the approach used in DatasetGAN~\cite{zhang2021datasetgan}, we employ Deeplab-V3~\cite{chen2017deeplab} with ResNet101~\cite{he2015deep} backbone as our 2D part segmentation network. 

\noindent\textbf{Experimental Setup.} Our baselines include transfer learning and DatasetGAN~\cite{zhang2021datasetgan}. For the transfer learning approach, we fine-tune the last layer of a pre-trained (on ImageNet) network with our human-annotated data.  
For the DatasetGAN baseline, we retrain a DatasetGAN model with the annotated data to generate a dataset comprised of 10K image-annotation pairs, which serves as the training set for the Deeplab-V3 segmentation network. We adjust the size of the concatenated feature to match that of the DatasetNeRF's feature size, ensuring a fair comparison.
For our DatasetNeRF, 
we generate 10K images with uniformly distributed angles which cover both frontal and horizontal views. The Deeplab-V3 segmentation network is then trained from scratch using this dataset. Each test dataset contains 90 unique individual image annotations for arbitrary poses within the training spectrum, as well as a sequential video annotation comprising 61 frames.

\noindent\textbf{Quantitative Evaluation.} 
\cref{tab:full_comparison} presents the segmentation results for the AFHQ-Cat and FFHQ datasets. DatasetNeRF outperforms the baseline models in terms of segmentation quality in the video test set on both AFHQ-Cat and FFHQ datasets. This enhancement underscores the informativeness of the data generated by our approach, which is attributed to the 3D consistency prior inherent from the pretrained generator backbone. Specifically, in the individual test set, DatasetNeRF achieves superior segmentation results for the AFHQ-Cat dataset and demonstrates comparable performance for the FFHQ dataset.
 
\noindent\textbf{Qualitative Evaluation for 3D-Consistency.}
We show spatiotemporal line textures \cite{Bolles1987EpipolarPlaneIA} of semantic masks from different poses in \cref{fig:spatiotemporal} in the supplementary material. The smoothness of these semantic textures, matching the corresponding RGB textures, demonstrates the 3D-consistency of our generated data. While modeling 3D-consistency in RGB space is challenging due to high-frequency details with 2D CNN upsampling module\cite{xiang2023gramhd}, semantic mask generation is comparably easier and performs well.

\begin{table}[ht]
\caption{Training results with different numbers of generated point cloud training samples on AFHQ-Cat dataset.}
\renewcommand\tabcolsep{35pt}
\centering
\begin{tabular}{lcc}
\toprule
Experiments & Accuracy  & mIoU \\
\midrule
400 (Generated)   & 0.8788 & 0.6268 \\
600 (Generated)  & 0.8809 & 0.6447 \\
800 (Generated)  & 0.8828 & 0.6403 \\
1100 (Generated) & \textbf{0.8950} & \textbf{0.6651} \\
\bottomrule
\end{tabular}
\label{tab:cat_point_cloud_training_results}
\end{table}

\begin{table}[htbp]
\caption{Evaluation of the generated point cloud on ShapeNet-Car dataset with PointNet as the backbone model.}
\renewcommand\tabcolsep{0.5pt}
\centering

\begin{tabular}{lcc}
\toprule
Experiments & Accuracy & mIoU \\
\midrule
600 (ShapeNet) & 0.8796 & 0.6773 \\
600 (ShapeNet) + 725 (Generated) & \textbf{0.9073} & \textbf{0.7519} \\
\midrule
1325 (ShapeNet) & 0.9059 & 0.7412 \\
1325 (ShapeNet) + 1325 (Generated) & \textbf{0.9104} & \textbf{0.7571} \\
\bottomrule
\end{tabular}

\label{tab:performance_comparison_shapenet_car}
\end{table}

\subsection{3D Point Cloud Part Segmentation}

We demonstrate the effectiveness of our generated point cloud part segmentation dataset by training PointNet~\cite{qi2017pointnet} on the generated data. We assess the performance on AFHQ-Cat faces and Shape-Net Car based on mean Intersection-over-Union (mIoU) and accuracy metrics.  We show that our approach not only enables the generation of high-quality new point cloud part segmentations dataset from self-annotated 2D images but also acts as a valuable augmentation to existing classical 3D point cloud part segmentation benchmark datasets.

\noindent\textbf{AFHQ-Cat Face Point Cloud Segmentation.} From 1200 generated samples, we create a fixed test set of 100 point clouds and train PointNet with varying numbers of training samples. \cref{tab:cat_point_cloud_training_results} illustrates that increased training samples enhance model performance on the test set, which shows the effectiveness of our self-generated point cloud part segmentation dataset. 

\noindent\textbf{Augmentation of ShapeNet-Car.} Our method's efficacy, detailed in \cref{tab:performance_comparison_shapenet_car}, shows its potential as both a substitute and an augmentation for the original ShapeNet-Car dataset in different experimental setups. 
The original benchmark dataset consists of 1,825 ShapeNet-Car point cloud part segmentations. Out of these, 500 point clouds are designated as the test set, and the remaining 1,325 serve as the training set. 

\begin{figure*}[ht]
    \centering
    \includegraphics[width=\textwidth]{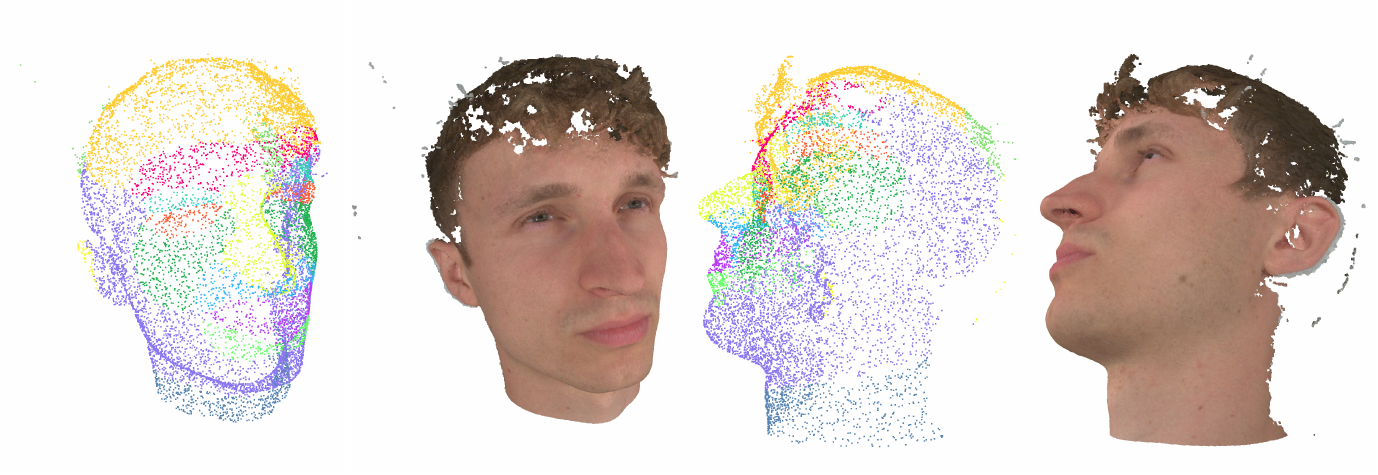}
    \caption{Visualization of real-world human face point cloud segmentations.
    }
    \label{fig:real_world_human_face_seg}
    
\end{figure*}

\noindent\textbf{Real-world Human Face Point Cloud Segmentation.}
We further demonstrate qualitatively that models trained on point clouds generated by our method exhibit effective generalization capabilities when applied to real-world datasets. We trained a PointNet segmentation model on a dataset comprising 1,200 point clouds, synthesized by our proposed method. For evaluation purposes, we employed the Nersemble dataset \cite{Kirschstein_2023}, extracting point clouds from 2D multi-view images as our test dataset. We show segmentation results for one sample in \cref{fig:real_world_human_face_seg}, showcasing the model's ability to achieve plausible segmentation on real-world human face, particularly in critical regions such as the neck, nose bridge, eyes, cheek, and forehead. Nonetheless, the model encounters challenges in correctly classifying certain areas, notably the ears, which are misidentified due to the model's limitations in distinguishing them from facial regions.

\begin{table}[htbp]
\caption{Ablation study comparing the impact of different settings on the dataset, focusing on mIoU and accuracy metrics.}
\renewcommand\tabcolsep{20pt}
    \centering
    \begin{tabular}{l|c|c}
        \toprule
        Settings & mIoU & Accuracy \\
        \midrule
        w/o Multiscale Feature & 0.4014 & 0.7067 \\
        w/ Multiscale Feature & \textbf{0.4796} & \textbf{0.7884} \\\hline
        w/o Density Prior & 0.4728 & 0.7813 \\
        w/ Density Prior & \textbf{0.4796} & \textbf{0.7884} \\
        \hline
         w/o Density Prior (Video) & 0.6899 & 0.9188 \\
        w/ Density Prior (Video) & \textbf{0.6913} & \textbf{0.9268} \\
       
        \bottomrule
    \end{tabular}
    \label{tab:ablation_study_triplane}
\end{table}

\begin{table}[htbp]
\caption{Ablation study on the effect of training sample size on mIoU and accuracy metrics for individual images and video sequences.}
\renewcommand\tabcolsep{15pt}
    \centering
    \begin{tabular}{lcccc}
        \toprule
        \multirow{2}{*}{Training Samples}
         & \multicolumn{2}{c}{AFHQ-Cat Individual} & \multicolumn{2}{c}{AFHQ-Cat Video} \\
        \cmidrule(lr){2-3} \cmidrule(lr){4-5}
         & mIoU & Acc. & mIoU & Acc. \\
        \midrule
        30 images & 0.4394 & 0.7716 & 0.6138 & 0.8892 \\
        45 images & 0.4588 & 0.7778 & 0.6752 & 0.9148 \\
        90 images & \textbf{0.4795} & \textbf{0.7884} & \textbf{0.6913} & \textbf{0.9268} \\
        \bottomrule
    \end{tabular}
    \label{tab:ablation_study_training_samples}
\end{table}

\begin{figure*}[ht]
  \centering
  \begin{minipage}[b]{0.49\textwidth}
    \includegraphics[width=\linewidth]{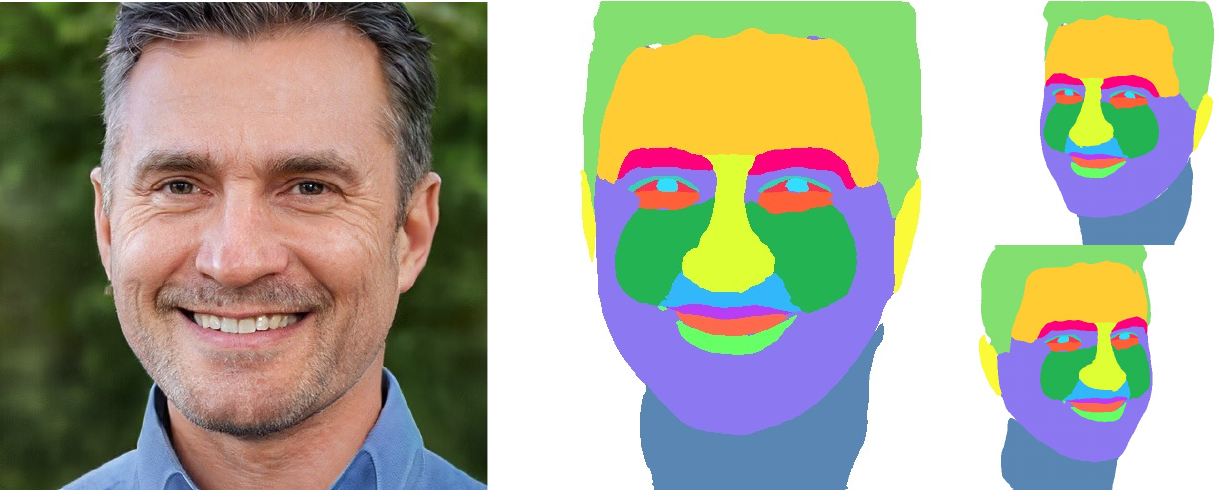}
  \end{minipage}
  \begin{minipage}[b]{0.49\textwidth}
    \includegraphics[width=\linewidth]{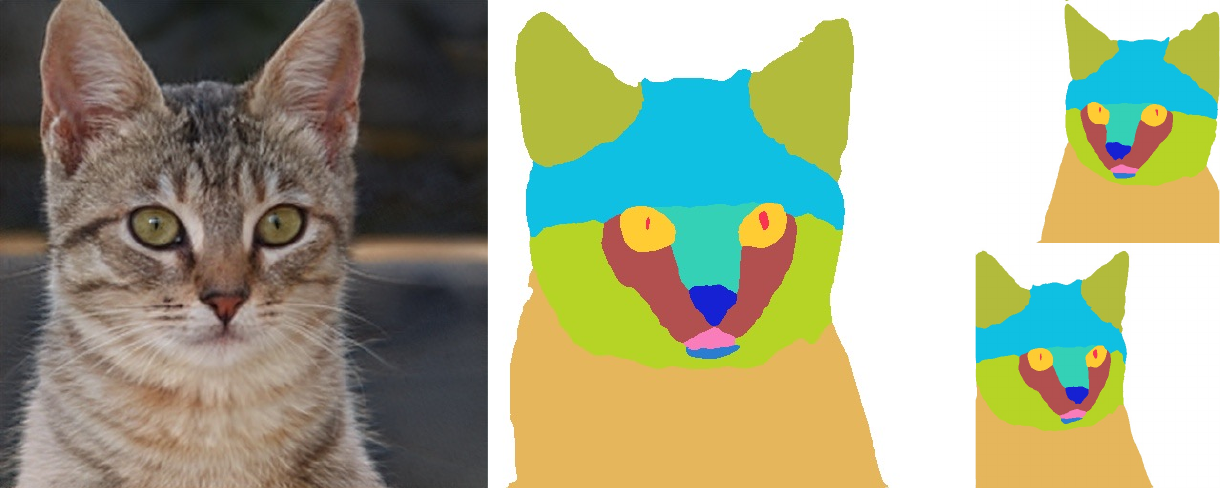}

  \end{minipage}

\caption{\textbf{3D RGB Inversion.} When presented with an arbitrarily posed input RGB image, our model concurrently optimizes the latent code \( \mathbf{z}\) and pose code to develop a 3D representation. It effectively functions as a segmentation model, capable of rendering segmentations from various viewpoints for the given input image.}

    \label{fig:rgb_inversion}
\end{figure*}

\begin{figure*}[ht]
    \centering
    \includegraphics[width=\textwidth]{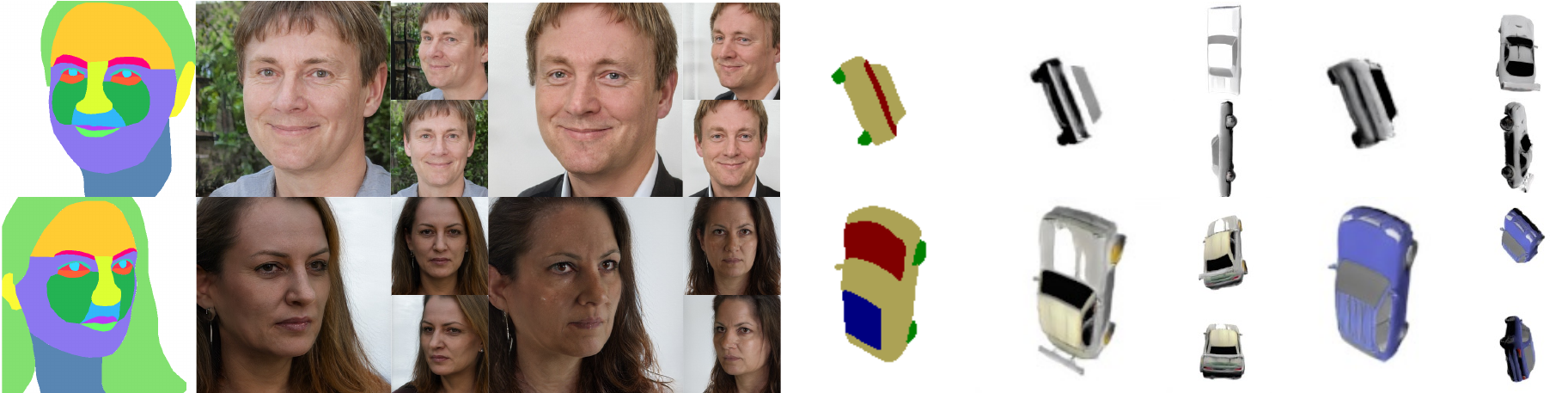}
    \caption{\textbf{3D Segmentation Inversion. } Given an arbitrary posed input semantic mask, we jointly optimize the latent code \(\mathbf{z}\)
 and pose code to construct a 3D representation.  The inherent 2D-to-3D ambiguity in this process results in a significant diversity in the 3D reconstructions.  This optimized representation allows rendering from various viewpoints.  }
    \label{fig:seg_inversion}
    
\end{figure*}

\subsection{Ablation Study}
\label{subsec:ablation_study}
In this section, we evaluate various aspects of our methodology. We ablate the experiments on AFHQ-Cat dataset. The testset is same as the testset used in \cref{sec:2dpartseg}. We employ GAN inversion~\cite{zhu2018generative} to initially optimize the latent code and pose of an input RGB test image, subsequently generating its corresponding semantic segmentation. We begin by examining the impact of the tri-plane architecture's size, as shown in \cref{tab:ablation_study_triplane}. Enhancing the original tri-plane architecture from EG3D with multiscale features extracted from the generator's backbone leads to a significant improvement in performance. Moreover, \cref{tab:ablation_study_triplane}  shows incorporating a density prior from the pretrained RGB decoder into the semantic branch is also beneficial.
Further, we investigate how the number of training samples affects performance in \cref{tab:ablation_study_training_samples}. Our findings suggest a moderate improvement when increasing the sample size from 30 to 90 images.

\begin{figure*}[htbp]
  \centering
  \begin{minipage}[b]{0.49\textwidth}
    \includegraphics[width=\linewidth]{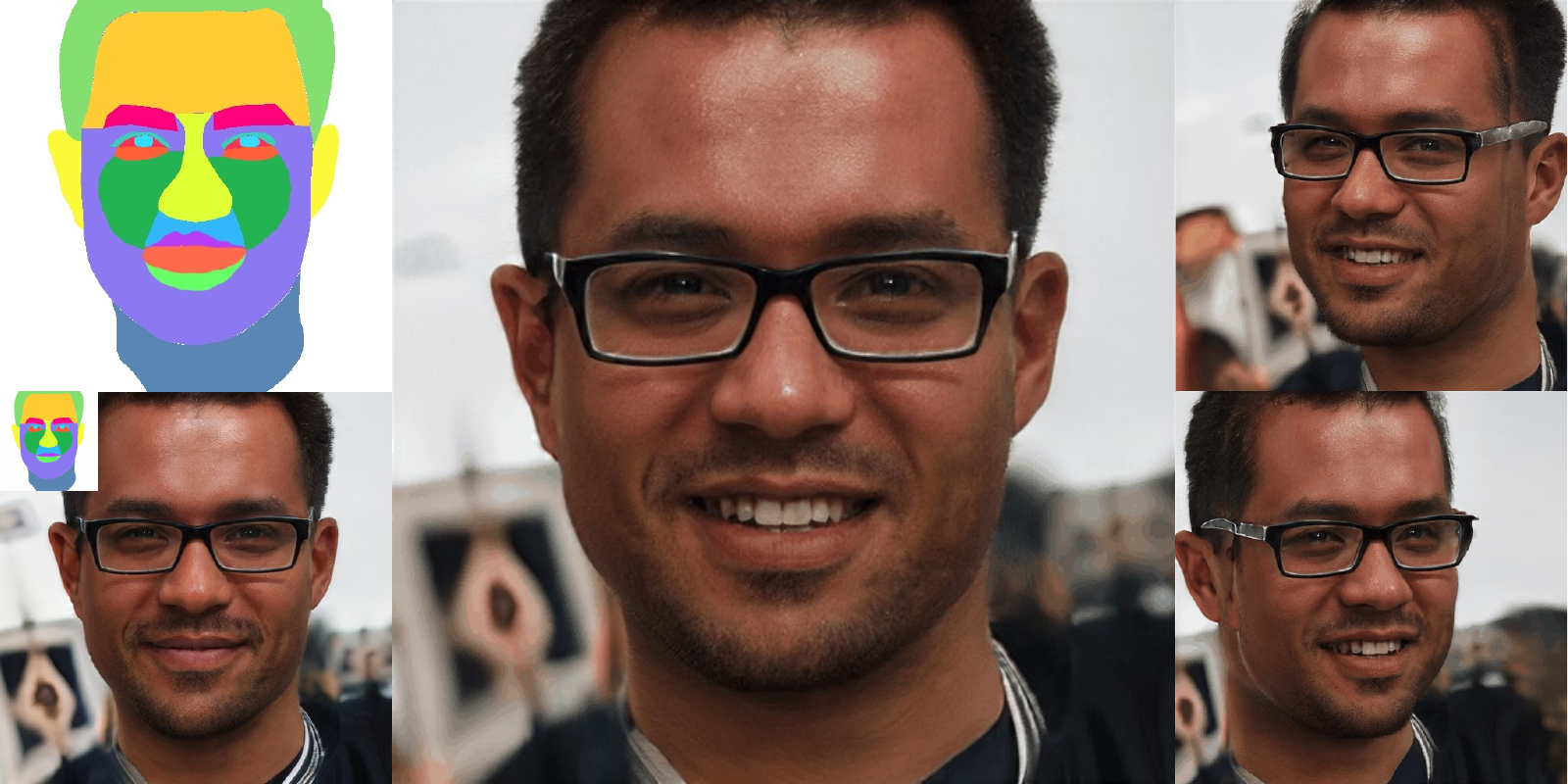}
  \end{minipage}
  \begin{minipage}[b]{0.49\textwidth}
    \includegraphics[width=\linewidth]{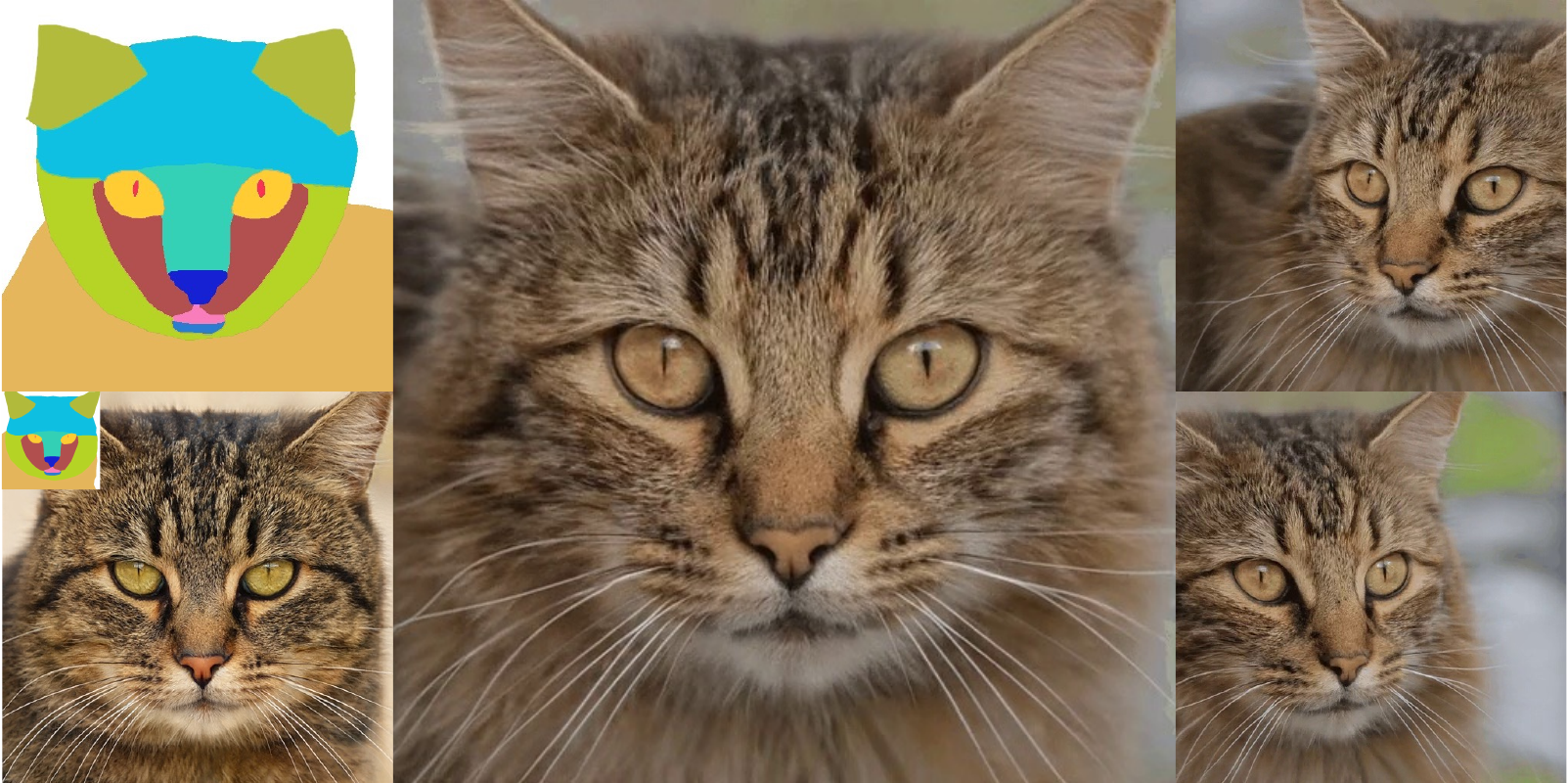}
  \end{minipage}
  \caption{\textbf{Semantic Editing Results.}  Our 3D editing system enables users to modify input label maps and subsequently acquire the corresponding updated 3D representation. We can render the updated 3D representation from different views.  }
    \label{fig:semantic editing}
  \end{figure*}
\subsection{Applications}

We explore a series of applications with our approach, including 3D inversion and 3D-aware editing.

\noindent \textbf{3D RGB Inversion.}  DatasetNeRF functions effectively as a segmentation model. When given a arbitrary posed RGB image, GAN inversion techniques~\cite{zhu2018generative} are employed to jointly optimize the input latent code and pose parameters. The optimized latent code uncovers the underlying 3D structure, thereby allowing for precise rendering of semantic segmentation from multiple viewpoints. The optimization is supervised by MSE 
loss and Adam~\cite{kingma2017adam} optimizer is used. The inversion result is showed in \cref{fig:rgb_inversion}.

\noindent \textbf{3D Segmentation Inversion.}  Pix2pix3D~\cite{deng20233d} introduces a conditional GAN framework to infer a 3D representation from an input semantic mask. While effective, this approach requires extensive training annotations and significant computational time. DatasetNeRF offers an alternative for accomplishing the similar task. Utilizing an arbitrarily posed semantic mask, our model conducts GAN inversion through its semantic branch. In this process, we jointly optimize the input latent code \(\mathbf{z}\)
 and the pose, employing cross-entropy loss and gradient descent as our optimization strategies. The Adam optimizer~\cite{kingma2017adam} is employed in this process.  The results of this process are illustrated in \cref{fig:seg_inversion}.
 
\noindent\textbf{3D-aware Semantic Editing.}  Our 3D editing system enables users to modify input label maps and subsequently acquire the corresponding updated 3D representation. To accomplish this task, our system focuses on updating the semantic mask output to align with the edited mask while preserving the object's texture through GAN inversion.  Initially, GAN inversion is employed to determine the initial latent code $\mathbf{z}$ from a given forward-oriented input image, which serves as the starting point for subsequent optimization, enhancing performance. 
Subsequently, this latent code is refined through GAN inversion to yield the optimized updated representation. We define the region of interest \(r\) as a binary mask which includes the union region of the label region before and after the edit. 
We define the loss function \( \mathcal{L}(\mathbf{z}; r) \) to quantify the quality of an edit based on the latent code \( \mathbf{z} \) and the region of interest \( r \). It is given by:
\begin{align*}
\mathcal{L}(\mathbf{z}; r) =\ & \lambda_1 \cdot \mathcal{L}_{\text{label}}(G_{\text{semantic}}(\mathbf{z}); M_{\text{edit}}) \nonumber \\
&+ \lambda_2 \cdot \mathcal{L}_{\text{rgb}}(\overline{r} \odot G_{\text{rgb}}(\mathbf{z}); \overline{r} \odot I_{\text{rgb}}) \nonumber \\
&+ \lambda_3 \cdot \mathcal{L}_{\text{vgg}}(\overline{r} \odot G_{\text{rgb}}(\mathbf{z}); \overline{r} \odot I_{\text{rgb}}),
\end{align*}

where:
\begin{itemize}
    \item \( G_{\text{semantic}}(\mathbf{z}) \) is the rendered semantic mask from $\mathbf{z}$ with the semantic branch $G_{semantic}$.
    \item \( G_{\text{rgb}}(\mathbf{z}) \) is the rendered RGB image with the RGB branch.
    \item \( M_{\text{edit}} \) is the edited semantic mask.
    \item \( \mathcal{L}_{\text{label}} \) is the cross-entropy loss for semantic consistency.
    \item \( \overline{r} \) is the complement of the region \( r \).
    \item \( \odot \) is the element-wise product.
    \item \( I_{\text{rgb}} \) is the original RGB image.
    \item \( \mathcal{L}_{\text{rgb}} \) measures the RGB prediction's mean squared error.
    \item \( \mathcal{L}_{\text{vgg}} \) is the perceptual loss calculated using a VGG-based network.
    \item \( \lambda_1, \lambda_2, \lambda_3 \) balance the loss components.
\end{itemize}

When editing on the FFHQ dataset, an additional identity loss~\cite{richardson2021encoding} is incorporated, which calculates the cosine similarity between the extracted features of both the input and edited faces.~\cref{fig:semantic editing} shows the edited results. 
\section{Limitations and Future Work}
One limitation of our approach is its reliance on the availability and suitability of pre-annotated data, which restricts its application to more general contexts, such as indoor scene segmentation. In addition, our current approach utilizes the generators of 3D GANs, such as EG3D\cite{chan2022efficient} and GNARF\cite{bergman2023generative}, as the backbones of our model. While these 3D GANs adeptly handle single-category data distributions, we aim in our future work to expand this approach to diffusion models, taking advantage of their broader generative diversity. Moreover, although reshaping the concatenated feature into a tri-plane structure has been shown to significantly enhance segmentation quality, it presents challenges in terms of memory efficiency and computational demands. An intriguing avenue for future research lies in identifying the most representative semantic features within the generator backbone, thereby optimizing memory usage and reducing computational load.

\section{Conclusions}

We present an efficient and powerful approach to developing a 3D-aware data factory, requiring only a minimal set of human annotations for training. Once trained, the model is capable of generating multi-view consistent annotations and point cloud part segmentations from a 3D representation by sampling in the latent space.
Our approach is versatile, compatible with both articulated 
and non-articulated generative radiance field models, making it applicable for a range of tasks such as consistent segmentation of human body poses.
This method facilitates advanced tasks like 3D-aware semantic editing, and 3D inversions including segmentation and RGB inversions. The capability of our model to efficiently produce a wide range of 3D-aware data from a limited set of 2D labels is not only crucial for training data-intensive models but also opens up new possibilities in various 2D and 3D application domains.

\section*{Acknowledgements}
Adam Kortylewski gratefully acknowledges support for his Emmy Noether Research Group, funded by the German Research Foundation (DFG) under Grant No. 468670075.


%
%
\bibliographystyle{splncs04}
\bibliography{main}

\clearpage 

\section*{Supplementary Material}

\renewcommand{\thefigure}{S\arabic{figure}}
\section{Self-annotated Fine-grained Dataset}
\begin{figure}[htbp]
  \centering
  \begin{minipage}[b]{0.49\textwidth}
    \includegraphics[width=\linewidth]{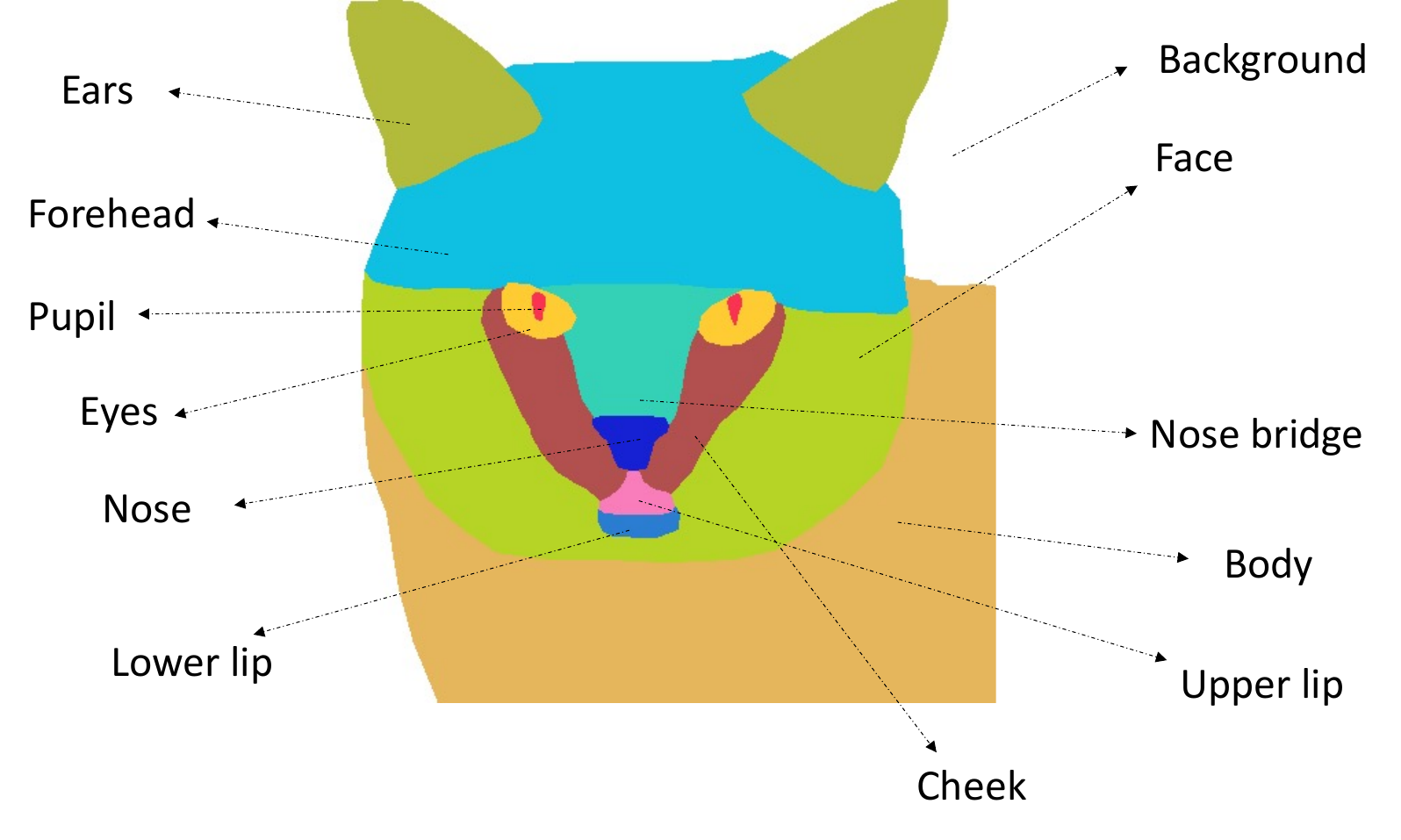}
  \end{minipage}
  \begin{minipage}[b]{0.49\textwidth}
    \includegraphics[width=\linewidth]{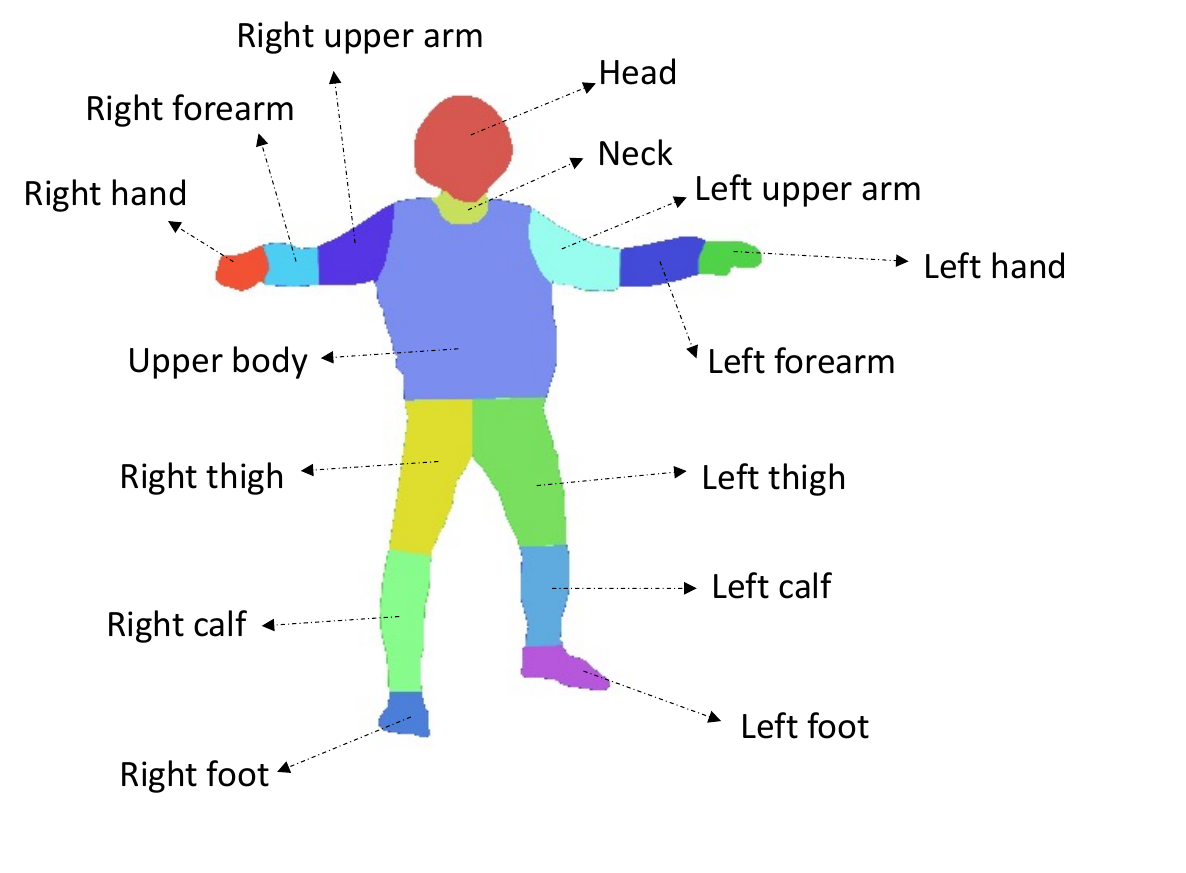}
  \end{minipage}
  \caption{Detailed label illustration for different datasets.}
  \label{fig: detailedannotation }
\end{figure}

We show more in detail about the fine-grained annotation dataset of AFHQ-Cat\cite{choi2020stargan}  and AIST++ dataset\cite{li2021ai} in Figure \ref{fig: detailedannotation  }.

\section{Analysis and Evaluation Metrics for 3D-Consistency} 
\begin{figure}[htbp]
    \centering
    \begin{subfigure}{0.47\linewidth}
        \includegraphics[width=\linewidth]{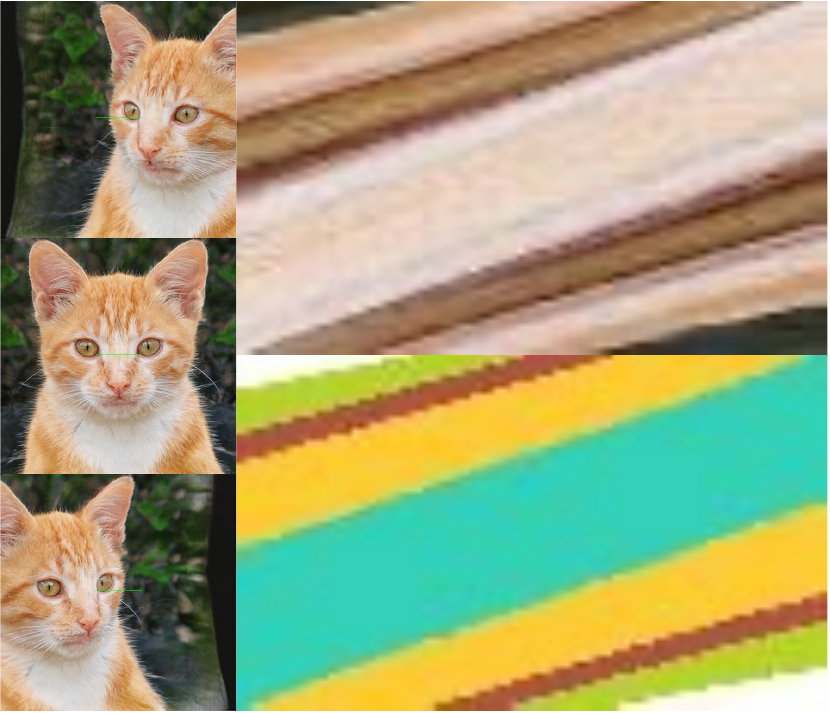}
        \caption{}
        \label{fig:spatiotemporal1}
    \end{subfigure}
    \hfill
    \begin{subfigure}{0.47\linewidth}
        \includegraphics[width=\linewidth]{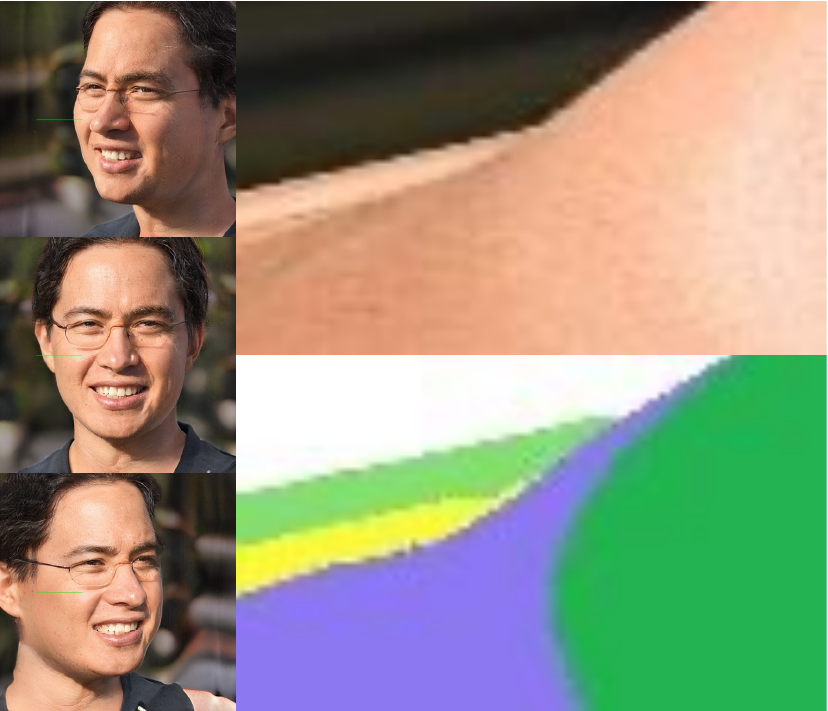}
        \caption{}
        \label{fig:spatiotemporal2}
    \end{subfigure}
    \caption{Spatiotemporal line RGB and semantic textures. Zoom in to better visualize the green line segments in the RGB images.}
    \label{fig:spatiotemporal}
   
\end{figure}

We show spatiotemporal line textures\cite{Bolles1987EpipolarPlaneIA} of semantic masks from different poses in \cref{fig:spatiotemporal}. 

\section{Real-world Human Face Scan Segmentation Results}
\begin{figure}[htb]
    \centering
    \begin{subfigure}[b]{0.45\textwidth} 
        \centering
        \includegraphics[width=\textwidth]{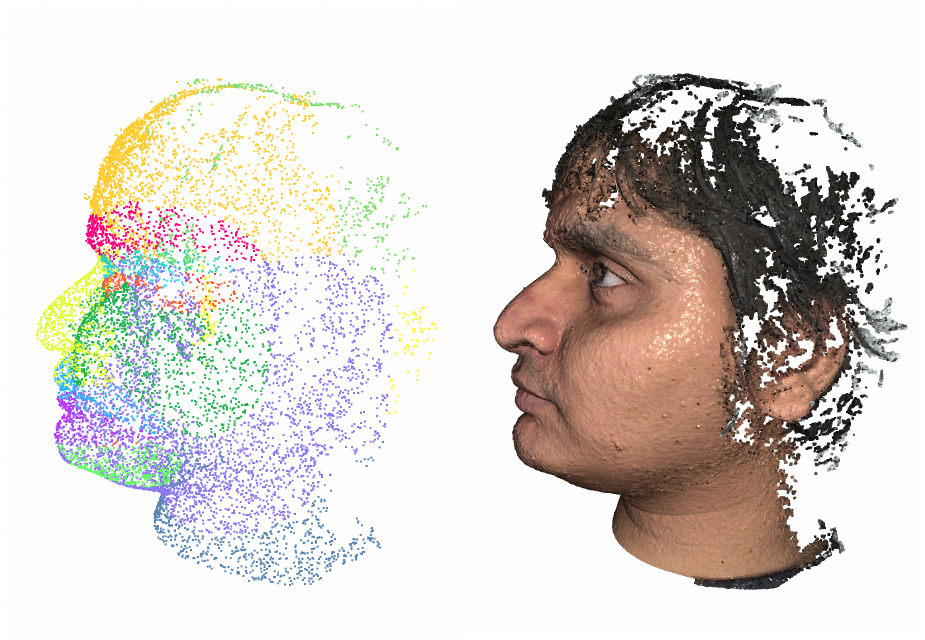}
        \caption{Scan segmentation result for the sample from Nersemble dataset\cite{Kirschstein_2023}. }
        \label{fig:left_subfig}
    \end{subfigure}
    \hfill 
    \begin{subfigure}[b]{0.45\textwidth} 
        \centering
        \includegraphics[width=\textwidth]{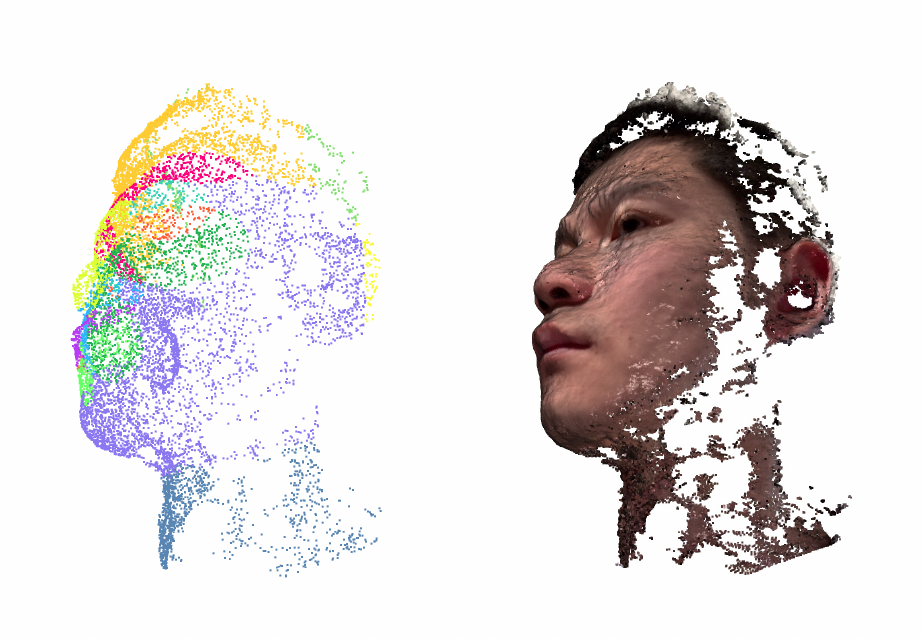}
        \caption{Scan segmentation result for the self-created scan. }
        \label{fig:right_subfig}
    \end{subfigure}
    \caption{Real-world human face scan segmentation results for samples from Nersemble dataset \cite{Kirschstein_2023} and self-made scan.}
    \label{fig:both_subfigs}
\end{figure}

We present additional real-world human face scan segmentation results. Figure \ref{fig:left_subfig} displays a sample selected from the Nersemble dataset\cite{Kirschstein_2023}, along with its segmentation outcomes. Conversely, Figure \ref{fig:right_subfig} shows a sample obtained from a self-scanned human face using a smartphone. Both examples demonstrate segmentation results that are both decent and reasonable.

\section{ShapeNet-Car Point Cloud Segmentation Performance Saturation} 
\begin{figure}[htbp]
  \centering
  \includegraphics[width=0.7\linewidth]{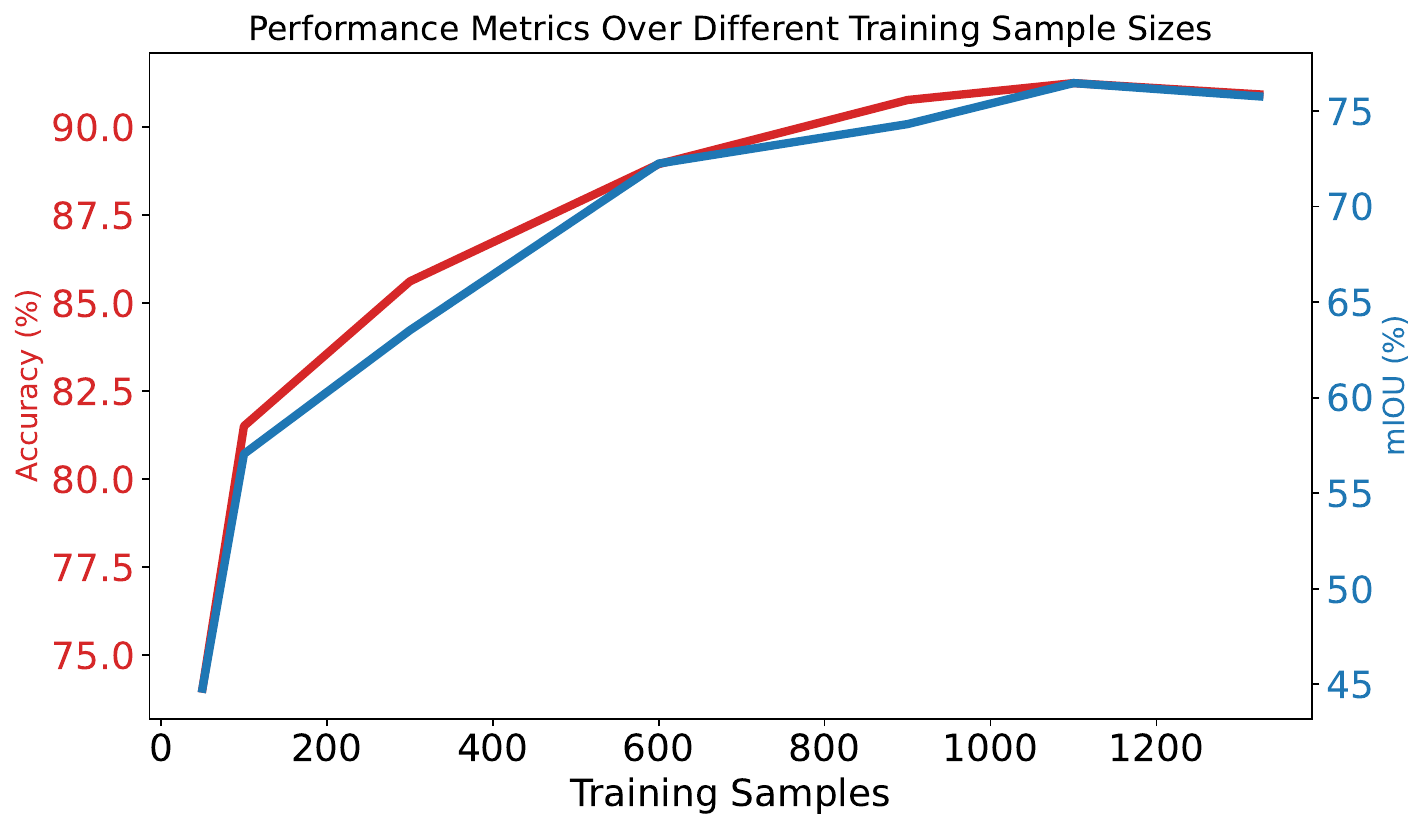}
  \caption{Experiments showing PointNet segmentation performance saturation. The red curve represents accuracy and blue curve represents mIoU.}
  \label{fig:saturation_shpaenet_car}
\end{figure}

ShapeNet-Car\cite{chang2015shapenet} Point Cloud Segmentation Performance Saturation: Table \cref{tab:performance_comparison_shapenet_car} in the main paper shows minimal performance gains due to a saturation in PointNet's capabilities, observed when training with approximately 1000 ShapeNet-Car samples (\cref{fig:saturation_shpaenet_car}). 

\section{Segmentation with Foundation Features} 
\begin{figure}[htbp]
  \centering
  \includegraphics[width=0.7\linewidth]{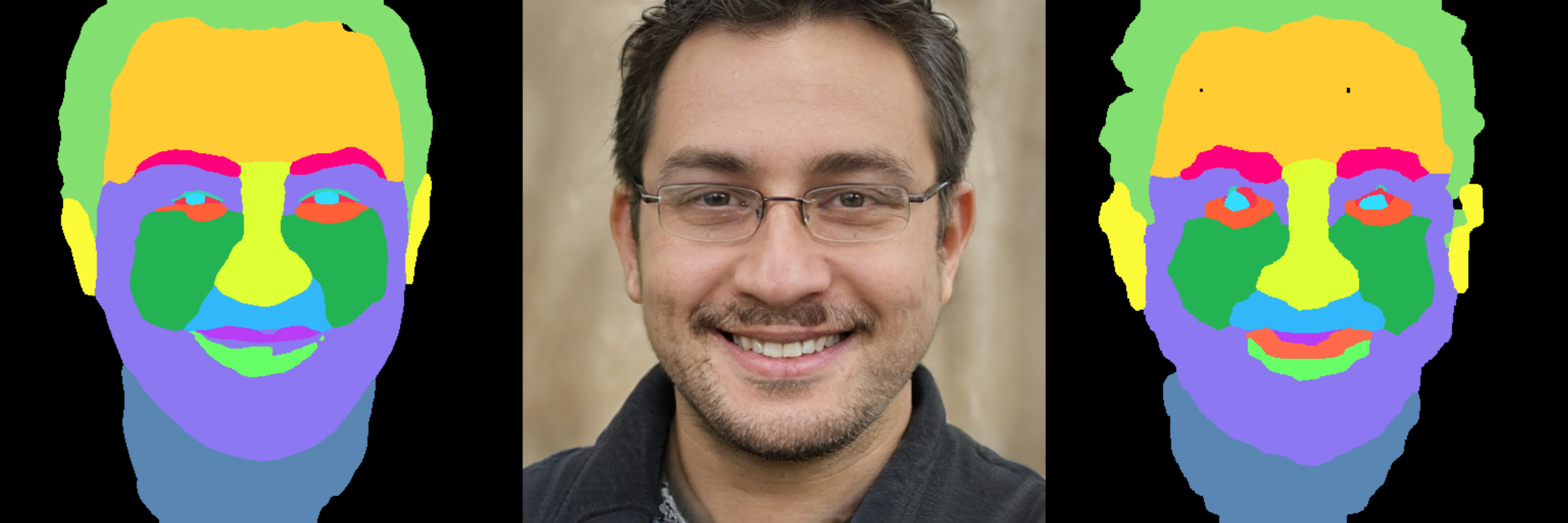}
  \caption{The comparisons with our result (left) with DINOv2 based method (right).}
  \label{fig:compare_with_dino}
\end{figure}

We conduct the suggested experiment and use pretrained DINOv2 features\cite{oquab2024dinov2} to train a segmentation model with our 90 annotated images. For testing, we feed the model frames from our test video sequences. As shown in \cref{fig:compare_with_dino}, our method's segmentation (left) is less noisy and more accurate than the DINOv2-based segmentation (right). 

\end{document}